\documentclass[10pt,twocolumn,letterpaper]{article}

\usepackage{3dv}
\usepackage{times}
\usepackage{epsfig}
\usepackage{graphicx}
\usepackage{amsmath}
\usepackage{amssymb}

\usepackage{comment}
\usepackage{color}
\usepackage{multirow}
\usepackage{arydshln}
\usepackage{cite}
\usepackage{subcaption}
\usepackage{floatrow}
\usepackage[pagebackref=true,breaklinks=true,colorlinks,bookmarks=false]{hyperref}

\usepackage{amssymb}
\usepackage{pifont}
\newcommand{\cmark}{\ding{51}}%
\newcommand{\xmark}{\ding{55}}%

\threedvfinalcopy 

\def\threedvPaperID{377} 

\ifthreedvfinal\fi

\newcommand{\highlightChange}{\color{red}}
\def\HC{\highlightChange}

\definecolor{amber_sae}{rgb}{1.0, 0.49, 0.0}
\definecolor{brandeisblue}{rgb}{0.0, 0.44, 1.0}
\definecolor{cadmiumorange}{rgb}{0.93, 0.53, 0.18}
\definecolor{carrotorange}{rgb}{0.93, 0.57, 0.13}
\definecolor{chocolate(web)}{rgb}{0.82, 0.41, 0.12}
\definecolor{cinnamon}{rgb}{0.82, 0.41, 0.12}	
\definecolor{cinnamon}{rgb}{0.82, 0.41, 0.12}
	\definecolor{cinnamon}{rgb}{0.82, 0.41, 0.12}
	\definecolor{mangotango}{rgb}{1.0, 0.51, 0.26}
	\definecolor{orange(colorwheel)}{rgb}{1.0, 0.5, 0.0}
	\definecolor{pumpkin}{rgb}{1.0, 0.46, 0.09}
	\definecolor{princetonorange}{rgb}{1.0, 0.56, 0.0}
	\definecolor{ruddybrown}{rgb}{0.73, 0.4, 0.16}
	\definecolor{safetyorange(blazeorange)}{rgb}{1.0, 0.4, 0.0}
	\definecolor{safetyorange(blazeorange)}{rgb}{1.0, 0.4, 0.0}
	\definecolor{tangelo}{rgb}{0.98, 0.3, 0.0}
	\definecolor{tenné(tawny)}{rgb}{0.8, 0.34, 0.0}
	\definecolor{blue(ncs)}{rgb}{0.0, 0.53, 0.74}
	\definecolor{cerulean}{rgb}{0.0, 0.48, 0.65}
	\definecolor{lapislazuli}{rgb}{0.15, 0.38, 0.61}
	\definecolor{mediumpersianblue}{rgb}{0.0, 0.4, 0.65}
	\definecolor{mediumtealblue}{rgb}{0.0, 0.33, 0.71}
	
\begin{document}


\title{
Joint 3D Human Shape Recovery and Pose Estimation from a Single Image \\
with Bilayer Graph \\
\hspace{10pt} \\
}


\author{Xin Yu\thanks{Work mainly done when Xin Yu was an intern at MERL.}\\
School of Computing, University of Utah \\ Salt Lake City, Utah, USA\\
{\tt\small xiny@cs.utah.edu}
\and
Jeroen van Baar\thanks{Corresponding author.}, Siheng Chen\\
Mitsubishi Electric Research Laboratories\\ Cambridge, MA, USA\\
{\tt\small \{jeroen, schen\}@merl.com}
}

\maketitle
\thispagestyle{empty}

\begin{abstract}
The ability to estimate the 3D human shape and pose from images can be useful in many contexts. Recent approaches have explored using graph convolutional networks and achieved promising results. The fact that the 3D shape is represented by a mesh, an undirected graph, makes graph convolutional networks a natural fit for this problem. However, graph convolutional networks have limited representation power. Information from nodes in the graph is passed to connected neighbors, and propagation of information requires successive graph convolutions. To overcome this limitation, we propose a dual-scale graph approach. We use a coarse graph, derived from a dense graph, to estimate the human's 3D pose, and the dense graph to estimate the 3D shape. Information in coarse graphs can be propagated over longer distances compared to dense graphs. In addition, information about pose can guide to recover local shape detail and vice versa. We recognize that the connection between coarse and dense is itself a graph, and introduce graph fusion blocks to exchange information between graphs with different scales. We train our model end-to-end and show that we can achieve state-of-the-art results for several evaluation datasets. The code is available at the following link,
\url{https://github.com/yuxwind/BiGraphBody}.

\end{abstract}

\section{Introduction}
\label{sec:intro}

\begin{figure}[ht!]
    \centering
     \begin{subfigure}[b]{0.39\textwidth}
         \centering
         \includegraphics[width=\textwidth]{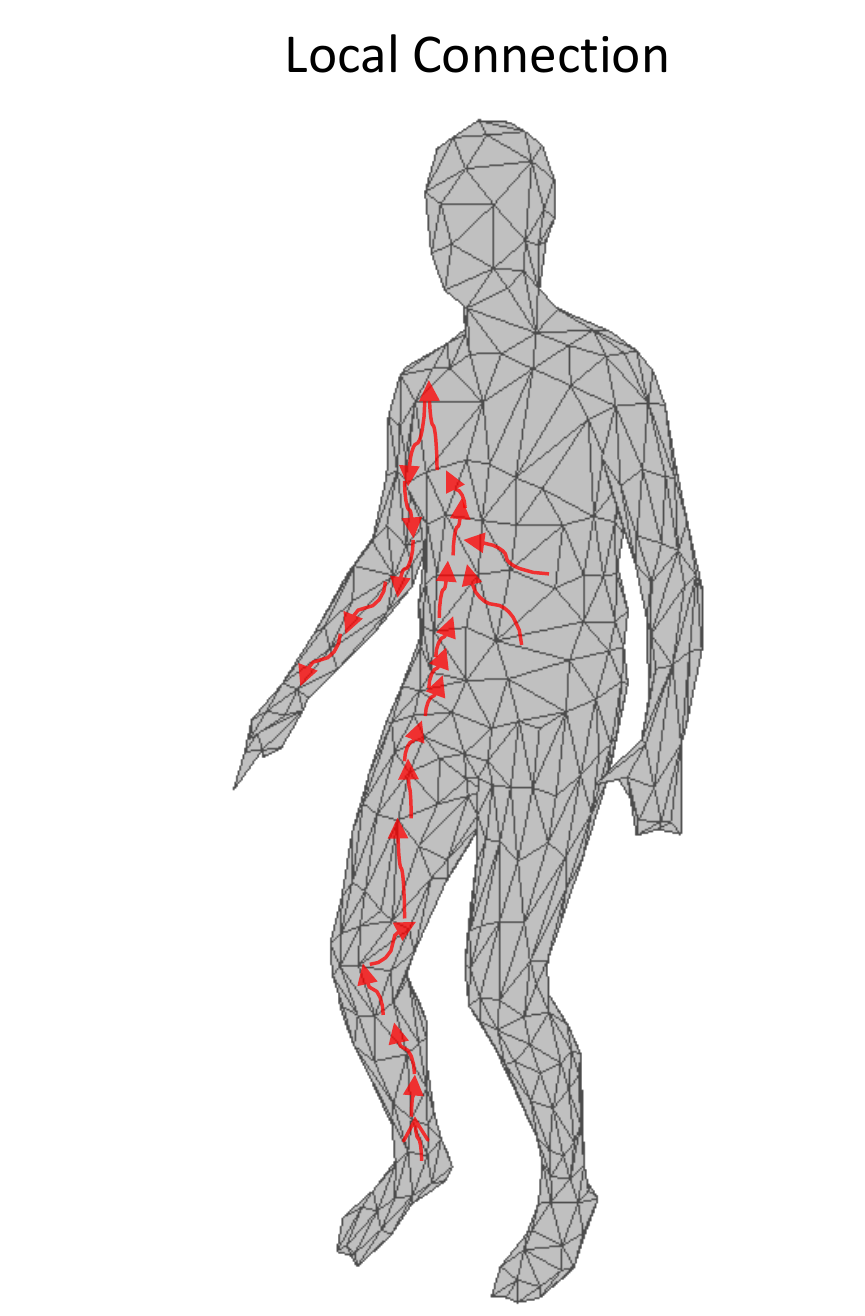}
         \caption{Mesh-only graph}
         \label{fig:local}
     \end{subfigure}
     \hfill
      \begin{subfigure}[b]{0.45\textwidth}
         \centering
         \includegraphics[width=\textwidth]{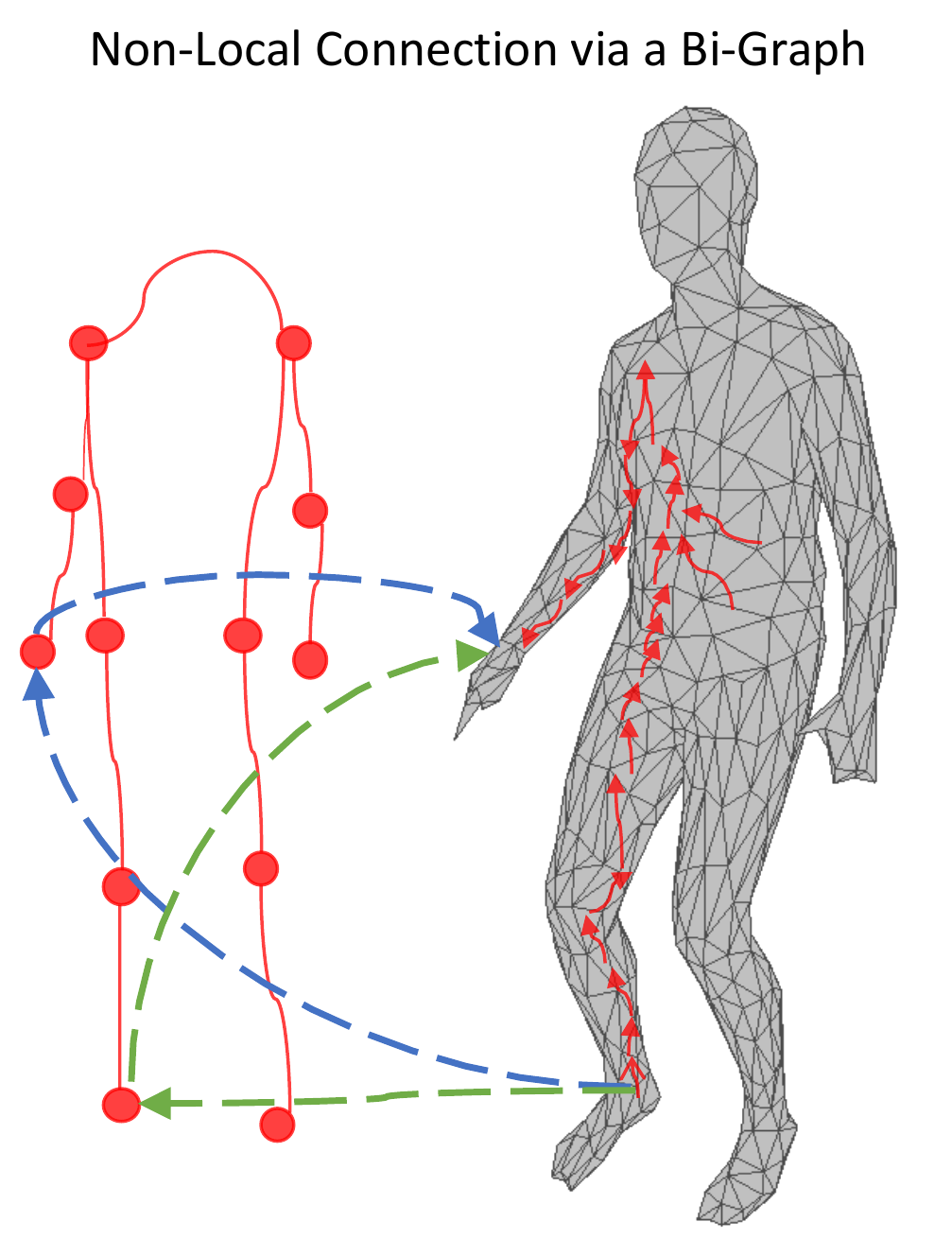}
         \caption{Bilayer graphs}
         \label{fig:non-local}
     \end{subfigure}
     \hfill
    \vspace{-2mm}
    \caption{By associating the mesh graph to the input image with a skeleton graph, the bilayer graph structure will shorten the paths between remote mesh nodes~(1723 nodes here), when we connect a joint with the mesh nodes it controls. With the body parts are correlated, such as the ankle and the wrist, this bilayer graph implicitly learn the interaction between joints and mesh vertices and further shorten the path among the remote body mesh vertices. 
    }
    \label{fig:motivation}
\end{figure}

Recovering 3D human shapes and poses from 2D images is a fundamental task for numerous real-world applications, such as animation and  dressing 3D people~\cite{bhatnagar2019multi, jiang2020bcnet}. Some recent approaches restrict themselves to only estimate 3D poses~\cite{Wandt_2019_CVPR, Zhang_2019_CVPR, Zhao_2019_CVPR}, while  other approaches need multiple images to  achieve reliable shape recovery~\cite{hmrKanazawa17, kolotouros2019cmr}. Here we consider joint 3D human shape recovery and pose estimation from a single image.

As an undirected graph, a 3D mesh can represent a human shape, making  graph-based techniques a natural fit to this task. For example, graph CMR~\cite{kolotouros2019cmr} deforms a template human mesh in a neutral pose to a desired shape through a graph convolutional network. 
Graph convolutional layers then propagate the node features over the mesh. However, this mesh-graph only approach suffers from the issue 
node feature propagation will be extremely slow when the mesh has dense vertices, such as 1723 nodes used for human in general. We illustrate this limitation of mesh based graph in Fig.~\ref{fig:motivation}.
The recent work ~\cite{lin2021end} use transformer to reduce the distance between any two nodes to 1 via self-attention mechanisms. However, self-attention over a down-sampled 423 mesh nodes is still not efficient; positional encoding may maintain the base coordinates information in the sequential ordering of the mesh nodes, still ignores the structured correlations between body parts.



To resolve the above issues, we propose a bilayer graph structure, where one layer is a mesh graph for human shapes, and the other layer is a newly added skeleton graph for body joints. As shown in Fig.~\ref{fig:motivation}, the newly added skeleton graph can associate the 2D body joints estimated from an image with their coordinates in 3D space.~\footnote{2D body joints can be well estimated from an image, e.g.,~\cite{cao2018openpose, sun2019deep}.} This body-joint-based correspondence allows us to attach detailed local image features to each body joint in the skeleton graph. 
In previous mesh-only graph approaches, as shown in Fig.~\ref{fig:local}, the ankle and wrist nodes in the mesh graph can only connect to each other via multiple iterations of aggregation-and-combine operations in GCN. Image feature propagation will be extremely slow when the mesh template has 1723 nodes. This bilayer graph structure (see Fig.~\ref{fig:non-local}) use sparse skeleton graph to guide the mesh nodes to exchange information in a more efficient way. It further shortens the paths between remote mesh nodes when connecting them via joints. We thus leverage the spatial non-locality of the mesh graph.

An added benefit of this two-layer graph structure is multi-tasking: achieving shape recovery and pose estimation at the same time. Two layers naturally model a human body from mesh and skeleton scales, handling shape recovery and pose estimation, respectively. The cross or fusion layer is a trainable bipartite graph that connects body joints and mesh nodes. Instead of imposing any fixed connections, such a bipartite graph can adaptively adjust the relationships between the mesh nodes and body joints. It enables the feature fusion between two scales of a human body, mutually enhancing two tasks. This is related to linear blend skinning~\cite{kavan2014part}, however, where skinning provides an analytical transformation, our cross layer learns a data-adaptive transformation between body joints and mesh nodes in the high-dimensional feature space.



In summary, our main contributions are:

$\bullet$ We are the first to propose a neural network based on a two-layer graph structure that jointly achieves 3D human shape and pose recovery. The skeleton graph module propagates pose (coarser-scale) information, the mesh graph module propagates detailed shape (finer-scale) information, and the fusion graph module allows us to exchange information across the two modules.
    
$\bullet$ We propose an adaptive graph fusion block to learn the  trainable correspondence between body joints and mesh nodes, promoting information exchange across two scales.

    
$\bullet$ We validate our method on several datasets (H36M, UP-3D, LSP), and show that exchanging local and global image information from different scales provides a significant improvement and speedup over single graph methods.  
\section{Related Work}
\label{sec:related}
\noindent{\bf Human Shape Recovery} Over the years there have been many approaches to recover 3D human shapes from images. Several methods propose to recover clothed humans from either  single or multi-view images~\cite{Alldieck_2019_ICCV, Gabeur_2019_ICCV, Pandey_2019_CVPR, Pumarola_2019_ICCV, pifuSHNMKL19, Yu_2019_CVPR, Zheng_2019_ICCV, Saito2020:PifuHD, jiang2020bcnet}. These approaches rely on well segmented humans in the images, and do not emphasize accurate 3D shape and pose. Our goal instead is to capture the 3D shape and pose accurately without relying on any prior segmentation. Other methods rely on the video input to recover 3D human shapes~\cite{Kanazawa_2019_CVPR, Pavlakos_2019_ICCV, Sun_2019_ICCV, Zhang_2019_ICCV}. While Our goal is to recover accurate 3D shape and pose from a single image only. To handle the alignment issue between neutral and deformed poses,~\cite{bogo2016keep} proposes an optimization procedure to iteratively refine the estimate. The authors in~\cite{Choi_2020, moon2020i2l} introduce a sequential and iterative approach from 2D poses. In this work, we propose a trainable two-layer graph structure to resolve the alignment issue which does not require iterations.


\begin{figure*}[t!]
    \centering
    \includegraphics[width=0.96\textwidth]{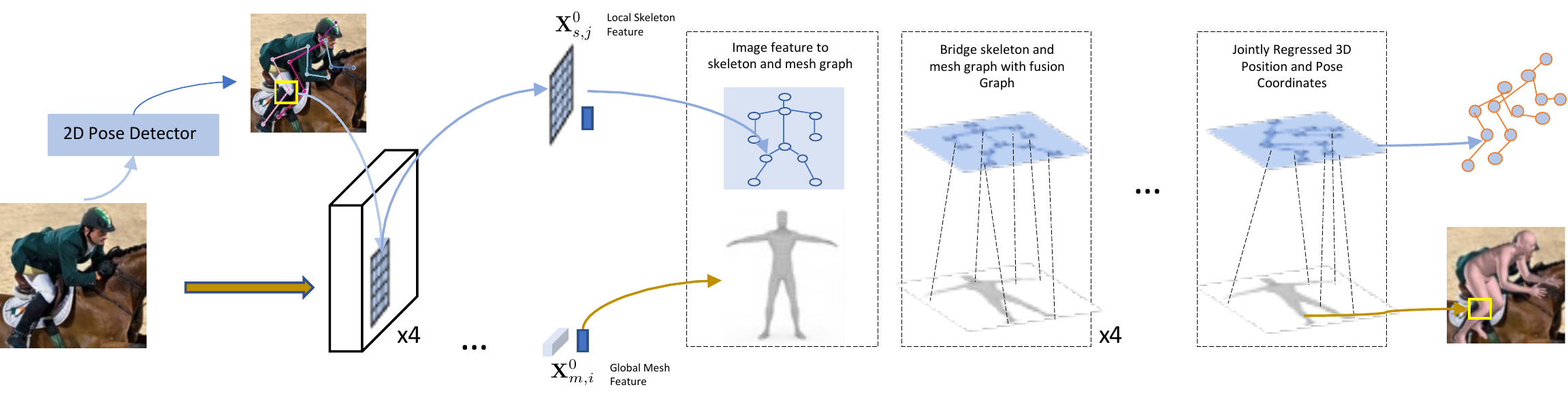}
    \vspace{-2mm}
    \caption{Our proposed bilayer graph architecture. Given an input image, the mesh graph module (Mesh-GCN) is a regression which outputs 3D vertex coordinates, and the skeleton graph module (Skeleton-GCN) estimates a skeleton with twelve 3D joint locations. The input to the Mesh-GCN is a template mesh together with a global perceptual feature extracted using a CNN, from the bounding box around the person in the image. Each global perceptual feature is attached to the XYZ coordinates of the vertices in the mesh. For clarity, we omit the SMPL part. The input to each joint node in the Skeleton-GCN is a local perceptual feature extracted from the image regions around the 2D joints estimated by HRNet. The two modules exchange information via so-called fusion graph, which is a bipartite graph between all mesh nodes and joints.}
    \label{fig:architecture}
    \vspace{-5pt}
\end{figure*}

\noindent {\bf Graph CNNs for 3D Reconstruction}
Recently, graph convolutional neural networks (GCNs) have been used to recover 3D objects from images. In this method, objects are represented as meshes~\cite{Wang_2018_ECCV, Wen_2019_ICCV}. Meshes are the de facto representation of 3D objects in computer graphics, and a mesh can be considered an undirected (3D) graph. The initial mesh before refinement may be a mesh obtained from a volumetric estimation~\cite{meshrcnn}. The authors in~\cite{Ci_2019_ICCV} state the limitations of GCN for 3D pose estimation. To overcome this limitation, they propose learnable weights for the structure of the graph. To address the limited representation capability of GCN, we propose a two-layer-graph neural networks with adaptive edge weights to share information between two graph layers. Our work is an extension to the regression based approach called Graph CMR~\cite{kolotouros2019cmr}. The input to Graph CMR is a human mesh in neutral pose along with global image features. Graph CMR then relies solely on graph convolutions to propagate information between nodes, and finally provide a 3D estimate. The mesh is refined by estimating SMPL parameters. We propose a two-layer graph to more efficiently propagate information. Multi-scale graphs have been explored in some other applications. The authors in~\cite{Gong_2019_CVPR} use multiple graph scales and exchange information via connectivity between different scales for the purpose of human parsing. The authors in~\cite{li2019symbiotic} use multi-scale graphs for the purpose of joint human action recognition and motion prediction. The information between scales is fused according to feature replication and concatenation between the different graphs. In comparison, our fusion approach relies on a learnable graph adjacency matrix, which exchanges information between two body scales.
\section{Problem Formulation}
As shown in Fig.~\ref{fig:motivation}, to resolve the issues of lacking detailed local information and inefficient long-range interactions, we use a bilayer graph structure to jointly estimate a 3D human pose and recover a complete 3D mesh based on a single input RGB image~(without knowing camera parameters). Mathematically, let 
$\mathbf{I} \in  \mathbb{R}^{H \times W \times 3}$ 
be an RGB image with the height $H$ and the width $W$. Both 3D human pose and 3D mesh structure can be represented as a graph with a set of node coordinates and an adjacency matrix indicating their connecting relations. For the 3D pose, we denote the skeleton pose coordinates as $\mathbf{V}_{\bf s} \in \mathbb{R}^{N_{\bf s} \times 3}$, where $N_s$ is the total number of body joints, thus the corresponding skeleton adjacency matrix is $\mathbf{A}_{\bf s} \in \mathbb{R}^{N_{\bf s} \times N_{\bf s}}$. For the 3D mesh structure, we denote the mesh node coordinates as  
$\mathbf{V}_{\bf m} \in \mathbb{R}^{N_{\bf m} \times 3}$, where $N_m$ is the number of mesh nodes. Then the adjacency matrix for the mesh structure is $\mathbf{A}_{\bf m} \in \mathbb{R}^{N_{\bf m} \times N_{\bf m}}$. We aim to propose a model 
$\mathcal{F}(\cdot)$: 
\begin{equation}
\widehat{\mathbf{V}}_{\bf s}, \widehat{\mathbf{V}}_{\bf m} = \mathcal{F}(\mathbf{I}, \mathbf{A}_m, \mathbf{A}_s),
\end{equation}
 to estimate human pose~$\widehat{\mathbf{V}}_{\bf s}$ and the recovered human mesh~ $\widehat{\mathbf{V}}_{\bf m}$, which precisely approximate the targets $\mathbf{V}_{\bf s}, \mathbf{V}_{\bf m}$, respectively. 

This joint task naturally requires to model a human body at two scales: a sparse graph at the skeleton scale and a dense graph at the mesh scale. To explicitly model the vertex correlations at two scales, we introduce a fusion graph to learn how the joints control the deformation of the body mesh vertices, and vice versa. Thus we propose a~\emph{two-layer graph structure} that consists of a skeleton graph $G_s(\mathcal{V}_s, {\bf A}_s)$, a mesh graph $G_m(\mathcal{V}_m, {\bf A}_m)$ and a fusion graph $G_f(\mathcal{V}_s \cup {V}_m, {\bf A}_f)$, where ${\bf A}_f \in \mathcal{R}^{(N_m+N_s) \times (N_m+N_s)}$ is the adjacency matrix for the fusion graph. Note that ${\bf A}_s,{\bf A}_m$ are fixed and given based on the human body prior; see the predefined graph topology in Fig.~\ref{fig:motivation};  while ${\bf A}_f$ is data-adaptive during training.

The graph-based formulation makes Graph CNN (GCN) a natural fit for this task. We propose a Bilayer-Graph GCN to address this task effectively and efficiently. As a core of the proposed system, it brings two benefits: First, it naturally models a human body from both mesh and skeleton aspects, promoting local and non-local topology learning, which will speedup the convergence of training and improve the joint pose and shape recovery. Second, a fusion graph enables information exchange between two scales of a human body, mutually enhancing feature extraction at two scales and further improving the performances in two tasks. 

\section{Two-scale Graph Neural Network}
\label{sec:technical}
To model the two-scale skeleton and mesh graph, and a fusion graph connecting them, we propose a bilayer graph neural network. Fig.~\ref{fig:architecture} shows an overview of the this architecture. In this section, we introduce the detailed implementation of each building block in our proposed method.


\subsection{Architecture Overview}
\label{subsec:arch-overview}
As shown in Fig.~\ref{fig:architecture}, given a single input image, an image encoder will be firstly used to extract the features from it, and a 2D-pose detector will processes the image into a skeleton graph. Then on the top part, skeleton graph module attaches local joint features, and propagate the features in skeleton graph layer. While at the bottom part, mesh graph module attach the global image features and models the mesh graph layer. Between them, fusion graph module connects between all the skeleton joints and mesh nodes, and exchange dual-scale information in a structured way. Finally, the learned joint and mesh node representation will be used to regress the 3D pose and mesh coordinates.

\subsection{Image Encoder}
\label{subsec: feature_extraction}
The functionality of an image encoder module is to extract informative visual features from an RGB image, which would be the input for the subsequent modules. Given an input RGB image $\mathbf{I}$,  we use a multi-layer CNN to obtain a collection of intermediate image features from the output of each layer $l$, $\{ \mathbf{X}^{(\ell)}_{\rm im} \}_{\ell=1}^{L} =  \mathcal{F}_{\rm im} \left( \mathbf{I} \right)$, where $\mathcal{F}_{\rm im}(\cdot)$ is a CNN whose architecture follows ResNet50~\cite{he2016deep}.~\footnote{Any ``typical'' CNN auto-encoder can be used for the image feature extraction.}

\subsection{Bilayer Graph Module} 
We propose to employ Graph CNN to jointly regress the 3D coordinates of the mesh and skeleton vertices. It consists of three sub-graphs: the mesh graph module, the skeleton graph and the fusion graph module. \\
For each sub-graph, we employ the same basic graph convolutions to formulate them ~\cite{Kipf:2016tc}, which is defined as:
\begin{equation}
\label{eq:graph_convolutions}
    \mathbf{X}^{\rm \ell+1} = {\bf A} \mathbf{X}^{\rm \ell}  \mathbf{W} \in \mathbb{R}^{N\times d_{\rm \ell+1}}
\end{equation}
where $\ell$ indicates the $\ell$-th convolutional layer, ${\bf A} \in \mathbb{R}^{N \times N}$ is a graph adjacency matrix for the (sub)graph, $\mathbf{W} \in \mathbb{R}^{d_{\rm \ell} \times d_{\rm \ell+1}}$ is a convolution weight matrix, $\mathbf{X}^{\rm \ell} \in \mathbb{R}^{N\times d_{\rm \ell}}$ is the input feature vector. As shown in Eq.~\eqref{eq:graph_convolutions}, given a set of vertices initialized with \textbf{input features}~($X^0$) and their \textbf{adjacency matrices}~($\bf A$), the graph convolution layer allows feature propagating and updating over the (sub)graph so that each vertex can aggregate information from its neighbors. In the following, we will introduce the functions of the three sub-graphs separately, and comparatively study the input and adjacency matrix of them.

\subsubsection{Mesh Graph Module}
The functionality of a mesh graph module is to regress the posed 3D mesh conditioned on the input image features. It is identical to graph layers of GraphCMR~\cite{kolotouros2019cmr}. We start from the template mesh in neutral (T-) pose introduced by SMPL and deform them to the shaped and posed mesh with the graph convolutions. 

\noindent\textbf{Inputs} We employ the template coordinates as the position embedding of the mesh vertices, and attach it with the 2048-D global feature vector of ResNet-50~\cite{he2016deep}  to feed in the mesh graph module. \\
Let $\mathbf{x}^{\rm g}_{\rm im} \in \mathbb{R}^{D_g}$ be the global image feature after the average pooling layer and 
$\mathbf{v}^{T}_{m,i} \in \mathbb{R}^3$ is the 3D coordinate of a $i$-th template mesh vertex. For each mesh vertex, we have  an initialized feature defined as: \\
\begin{equation}
\label{eq:mesh_feature}
\mathbf{X}^0_{m,i} = \mathcal{F}^{0}_{m}(\mathbf{v}^T_{m,i} \oplus \mathbf{x}^{\rm g}_{\rm im})  \in \mathbb{R}^{d_0}  
\end{equation}
Where $\oplus$ denotes feature vector concatenation and $\mathcal{F}_{m}^0$ denotes the linear layer to reduce the dimension (the typical reduced dimension is 512) of the concatenated features whose weights shared among all mesh vertices.\\
\noindent\textbf{Mesh adjacency matrix} $A_m$ is initialed as a binary matrix to indicate the connectivity among the vertices as shown in Fig.~\ref{fig:motivation} and further row-normalized. 


\subsubsection{Skeleton Graph Module}
\label{susec:skeleton}
The functionality of a skeleton graph module is to lift the 2D pose estimated from an input RBG image to a 3D pose.
As shown in Fig.~\ref{fig:non-local}, the sparse skeleton graph can promote the non-local topology features of the dense mesh graph and enhance the correlations between different body parts. Furthermore, we extracted the local features around the joints for precise pose.  


\noindent\textbf{Inputs} Instead of using global image features for all the joint nodes,  we use joint-aware local features for joint nodes. Given the image, we use HRNet~\cite{sun2019deep} off-the-shelve to estimate the 2D positions of body joints in this image. For each body joint, we crop a patch centered at the estimated 2D positions with the size of the average estimated bone length from the joints per image, using RoI Align~\cite{he_mask_rcnn} from the $k$-th image feature map~($K$ layers in total) $\mathbf{x}_{\rm im}^{k}$ of ResNet-50~\cite{he2016deep}. This feature patch reflects the local visual information around the corresponding body joint. We concatenate the image feature patches with the positional embedding as the initial skeleton features as:
\begin{equation}
\mathbf{X}^0_{s,i} = \mathcal{F}^{0}_{s}(\hat{\mathbf{v}}_{s,i} \oplus {\rm \mathbf{RoI}}(\hat{\mathbf{v}}_{s,i}, \mathbf{x}_{\rm im}^{1}, \dots \mathbf{x}_{\rm im}^{K}))
\in \mathbb{R}^{d_0}
\end{equation}
where $i$-th body joint estimated by HRNet as $\widehat{\mathbf{s}}_i \in \mathbb{R}^2$,  $\mathbf{RoI}(\cdot)$ returns image feature patches from $\mathbf{x}_{\rm im}^{k}$ using RoI Align~\cite{he_mask_rcnn} with the patch centered at $\widehat{\mathbf{s}}_i$. Similarly, $\mathcal{F}^{0}_{s}$ is a linear layer and share the weights among the skeleton vertices.
We also experimented with the skeleton template coordinates as the positional embedding but we did not observe quantitative improvement in the results and thus keep the 2D embedding for all experiments. 

\noindent\textbf{Skeleton adjacency matrix}
We use fixed adjacency matrix for ${\bf A}_s$. The element is initialized as the reciprocal of the Euclidean distance between two template joint vertices. 



\subsubsection{Fusion Graph Module}
\label{subsec:graph_fusion_blocks}
The functionality of a fusion graph is to correlate the sparse skeleton graph and the dense mesh graph and enable information exchange between them and mutually enhance both tasks of 3D shape recovery and pose estimation. As shown in Fig.~\ref{fig:non-local}, the fusion graph connection can shorten the path of two remote mesh vertices dramatically and will speedup the non-local information propagation of the mesh graph. 

\noindent\textbf{Inputs} The fusion graph consists of the vertex from both the Mesh Graph and Skeleton Graph and applies the same initial feature as those two sub-graphs. The intermediate input of this module are the intermediate features from the Skeleton Graph and Mesh Graph, which fuses those intermediate features and populates them back to both Skeleton-GCN and Mesh-GCN.\\
\noindent\textbf{Fusion adjacency matrix} To fuse features from the skeleton and mesh graph, we leverage a trainable fusion graph to reflect the data-driven connectivity between body joints and mesh nodes. Besides defining a fixed connection part, denoted as $A_{f,s}$, we allow an extra dynamic connection, denoted as $W_f$, to be trainable to capture the connectivity in the hidden feature space in a data-driven manner. We define the final adjacency matrix as:
\begin{equation}
\label{eq:Af}
    A_{f} = {\rm RowNorm}(A_{f,s} \odot W_f)
\end{equation}
where $A_{f,s}, W_f \in \mathcal{R}^{(N_m+N_s) \times (N_m+N_s)}$, $\rm RowNorm()$ indicate a row normalization, $\odot$ denotes element wise product. $W_f$  is learnable and its element is initialized as 1 for
vertex-joint correlation and 0 for joint-joint and vertex-vertex (the Skeleton and Mesh Graphs have cover those connections). The element of $A_{f,s}$ for a connection between a skeleton vertex and mesh vertex is fixed to the reciprocal of their Euclidean distance; otherwise, it is zero. 

We also experiment the $\rm RowNorm()$ with a softmax on each row and apply additive optation between $A_{f_s}$ and $W_f$, both of which bring minor change to the performance. We adopt the row normalization and element-wise product for their simplification of computation.

\subsubsection{Architecture Implementation}
\begin{figure}[t]
    \centering
    \includegraphics[width=0.7\linewidth]{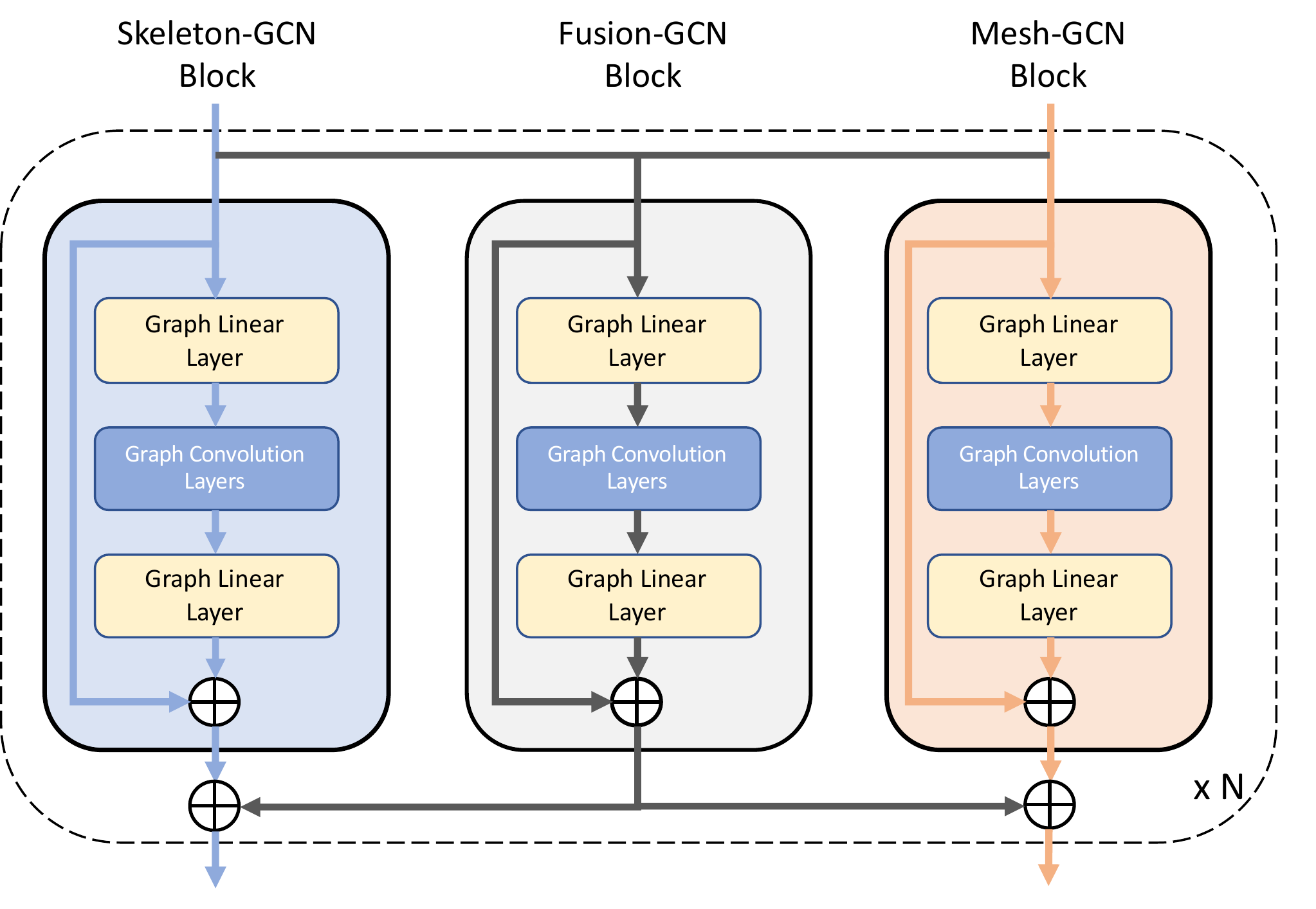}
    \vspace{-3.5mm}
    \caption{A Bilayer-Graph Block consists of a Fusion-GCN block, a Mesh-GCN block and a Fusion-GCN block. The fusion block depicted here has its input from the previous Skeleton-GCN and Mesh-GCN blocks prior to this Bilayer-Graph block, and adds its output back to each branch after their respective blocks.}
    \label{fig:graph_fusion_block}
\end{figure}

\noindent\textbf{Bilayer-Graph Block} At the heart of this approach, we propose a Bilayer-Graph block as an elementary computational unit for feature learning and propagation based on the bilayer-graph structure. As illustrated in Fig.~\ref{fig:graph_fusion_block}, the Skeleton-GCN Block, Fusion-GCN Block and Mesh-GCN Block apply graph convolutions on skeleton graph, fusion graph and mesh graph respectively. The Fusion-GCN Block collects features from both Mesh-GCN and Skeleton-GCN and distributes the updated features back to the other module. Each block consists of a sequence of a graph linear layer, a graph convolution layer, and another graph linear layer, with a residual connection from the input directly to the output of this block. Each layer follows a group normalization layer~\cite{Wu_2018_ECCV} and ReLU. Please note that the graph linear layer is a special graph convolution layer, which simply substitutes the graph adjacency matrix $ {\bf A}$ in the graph convolution layer ( see Eq. ~\eqref{eq:graph_convolutions}) to an identity matrix. 

In this network, we stack five Bilayer-Graph blocks for feature propagation, followed by two graph linear layers to regress the skeleton vertices and joint vertices separately. The first linear layers also follows a group normalization~\cite{Wu_2018_ECCV} and ReLU.\\

\noindent\textbf{SMPL regressor} As the parametric representation of the human body can be very useful for down-stream tasks (e.g., body manipulation), we follow~\cite{kolotouros2019cmr} to train a MLP module to regress pose ($\widehat{\theta}$) and shape ($\widehat{\beta}$) parameters for a SMPL model~\cite{SMPL:2015} from the predicted mesh $\widehat{\mathbf{V}}_m$.


\subsection{Training}
\label{subsec:loss_func}
\noindent\textbf{Losses} 
To train the Bilayer GCN, we apply loss functions on the output of the Bilayer GCN and SMPL regressor and minimize the errors between the predictions and ground truths. Firstly, we use the a per-vertex $L_1$ loss between the ground truth $\mathbf{V}_m$ and predicted mesh vertices $\hat{\mathbf{V}}_{m}$ from Mesh-GCN, denoted as $\mathcal{L}_m$, and between the GT and predicted joint vertices $\hat{\mathbf{V}}_s$ from Skeleton-GCN, denoted as $\mathcal{L}_s$. 

We follow ~\cite{kolotouros2019cmr, Kanazawa_2018_CVPR} to multiply the predicted mesh $\hat{\mathbf{V}}_m$ by a predefined matrix to get 3D joints, denoted as $\hat{\mathbf{V}}^{j3d}_m$. $L_1$ loss is also applied to it and its GT $\mathbf{V}_s$, denoted as $\mathcal{L}^{j3d}_m$. 

As we trained on mixed datasets consisting of both 3D and 2D data, we have additional supervision on the predict a weak perspective camera parameters from the intermediate features of the Mesh-GCN with two graph linear layers. Apply this camera parameters to $\hat{\mathbf{V}}^{j3d}_m$ and $\hat{\mathbf{V}}_s$, we get two sets of 2D pose and use a $\mathcal{L}_1$ loss on them and the 2D GT pose, denoted as $\mathcal{J}^{j2d}_m$ and $\mathcal{J}^{j2d}_s$ respectively.

Finally, we apply MSE loss on the predicted SMPL shape~($\hat{\theta}$) and pose~($\hat{\beta}$) parameters, denoted as $\mathcal{L}_{\theta}$ and $\mathcal{L}_{\beta}$ respectively. And we have the final loss as below:
\begin{equation}
\label{loss_function}
    \mathcal{L} =  \mathcal{L}_{\rm m} + \mathcal{L}^{j3d}_m + \mathcal{L}_{m}^{j2d} + \mathcal{L}_s + \mathcal{L}_{s}^{j2d} 
    + \mathcal{L}_{\theta} + \lambda\mathcal{L}_{\beta},
\end{equation}
\noindent\textbf{Focal loss for regression} We observe that in the above losses on 3D vertices, the error caused by each body part varies a lot. For example, the joints on legs and arms usually have much larger error than the other parts. The intuition is that the variation for body limbs is much larger compared to torso and head. We generalizes the focal loss~\cite{2017FocalLF}, which addresses class imbalance by down-weighting the loss for well-classified samples,  to this regression tasks to addresses the imbalanced vertex error. 
We modify it based on the $L_1$ loss of the target, i.e.,
$$\mathcal{L}_{fl} = -(\alpha \mathcal{L})^{\gamma} \log(1-{\rm max}(\tau, \alpha \mathcal{L})),$$
where $\mathcal{L}$ is the $L$1 loss, $\alpha$ is a factor to scale $\mathcal{L}$ to (0,1), $\tau < 1$ is a threshold that truncates $\alpha\mathcal{L}$ with a maximum value to avoid unreasonably large loss when $\alpha \mathcal{L}$ approaches 1, $(\alpha \mathcal{L})^{\gamma}$ is a factor to reduce the relative loss for well-regressed vertices with $\gamma > 0$.

\begin{figure*}[!t]
\centering
    \begin{subfigure}{.25\textwidth}
        \includegraphics[width=0.99\linewidth]{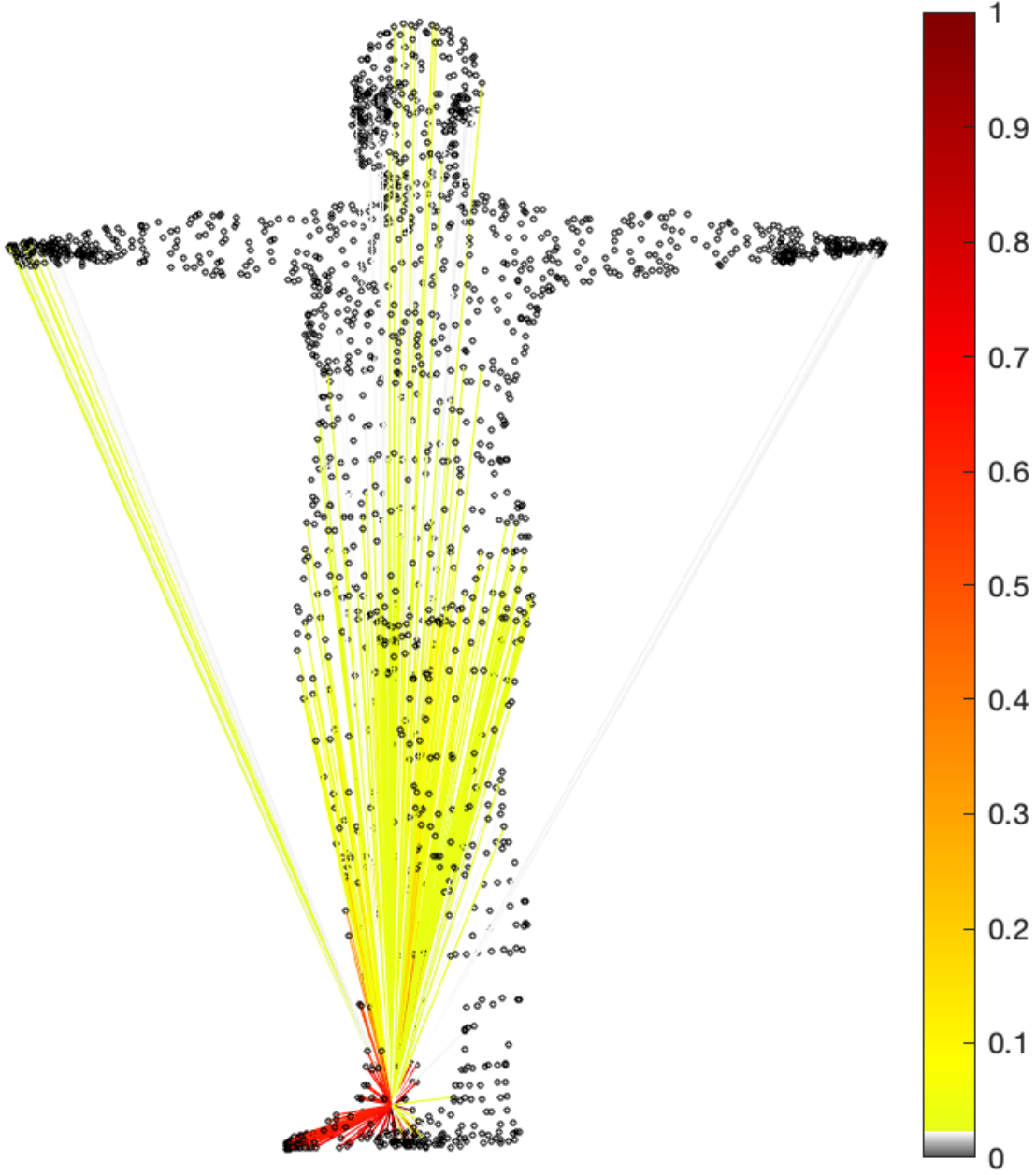}
    \end{subfigure}
    \hspace{6mm}
    \begin{subfigure}{.12\textwidth}
        \includegraphics[width=0.99\linewidth]{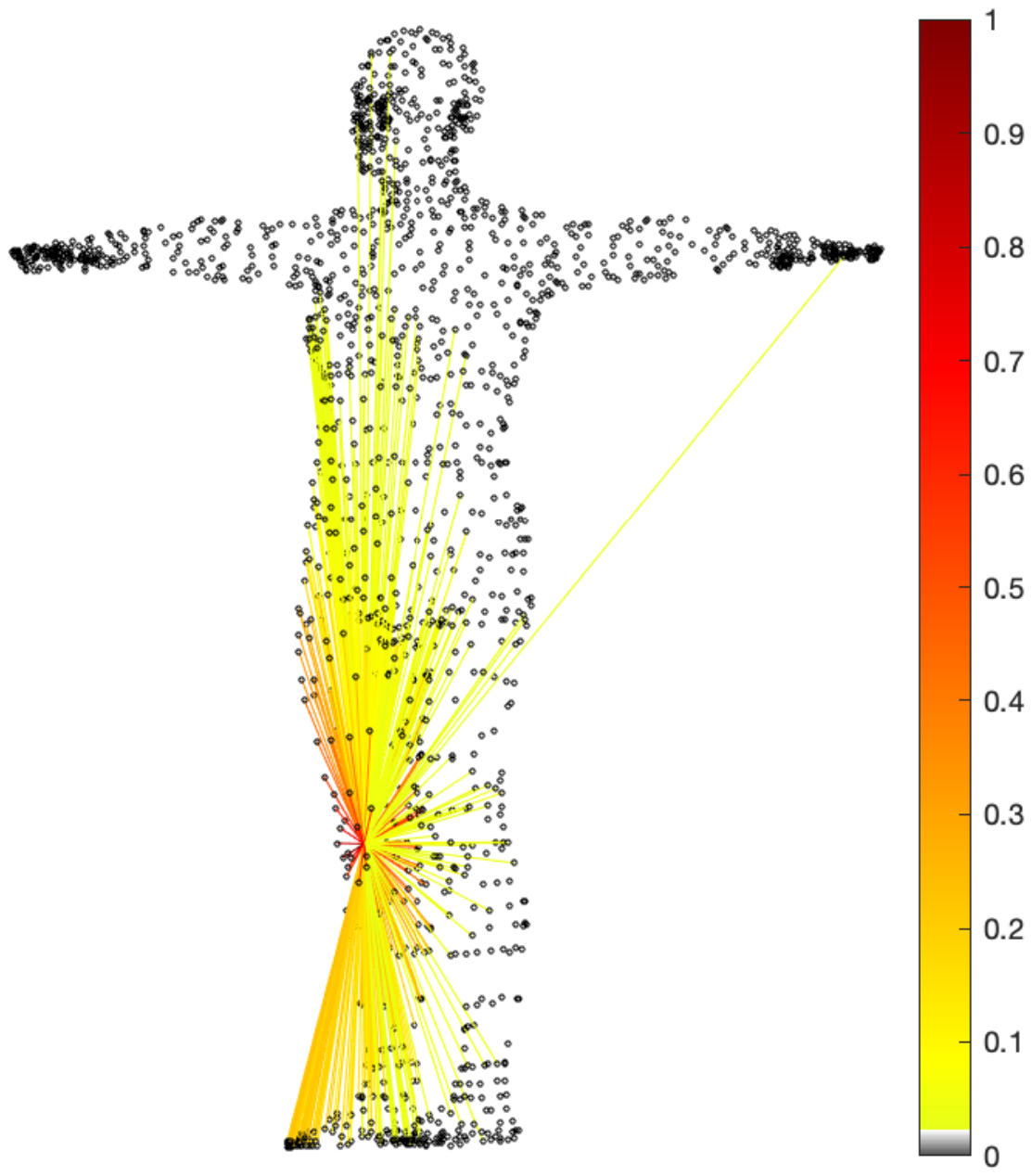}
        \includegraphics[width=0.99\linewidth]{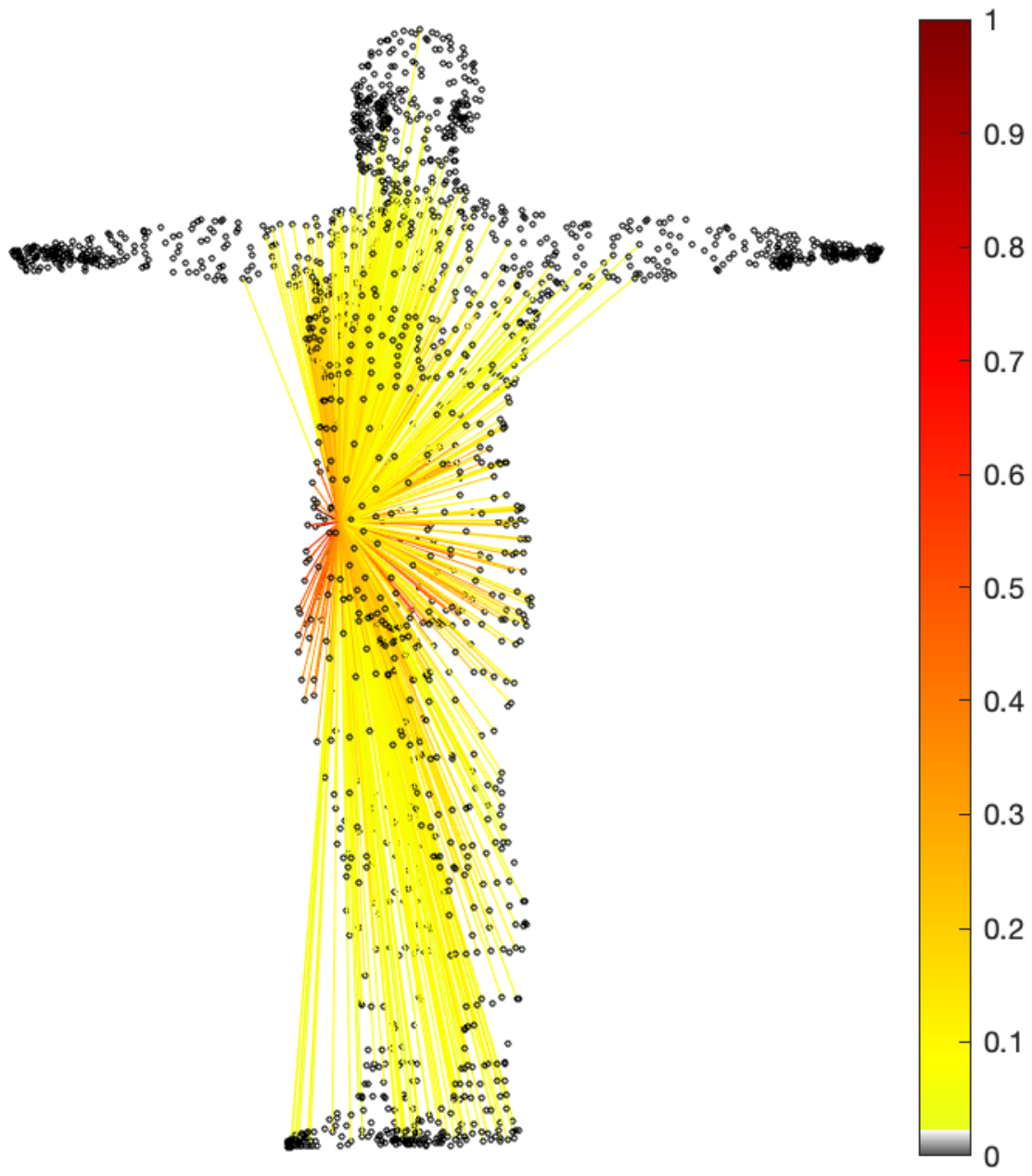}
    \end{subfigure}
    \begin{subfigure}{.12\textwidth}
         \includegraphics[width=0.99\linewidth]{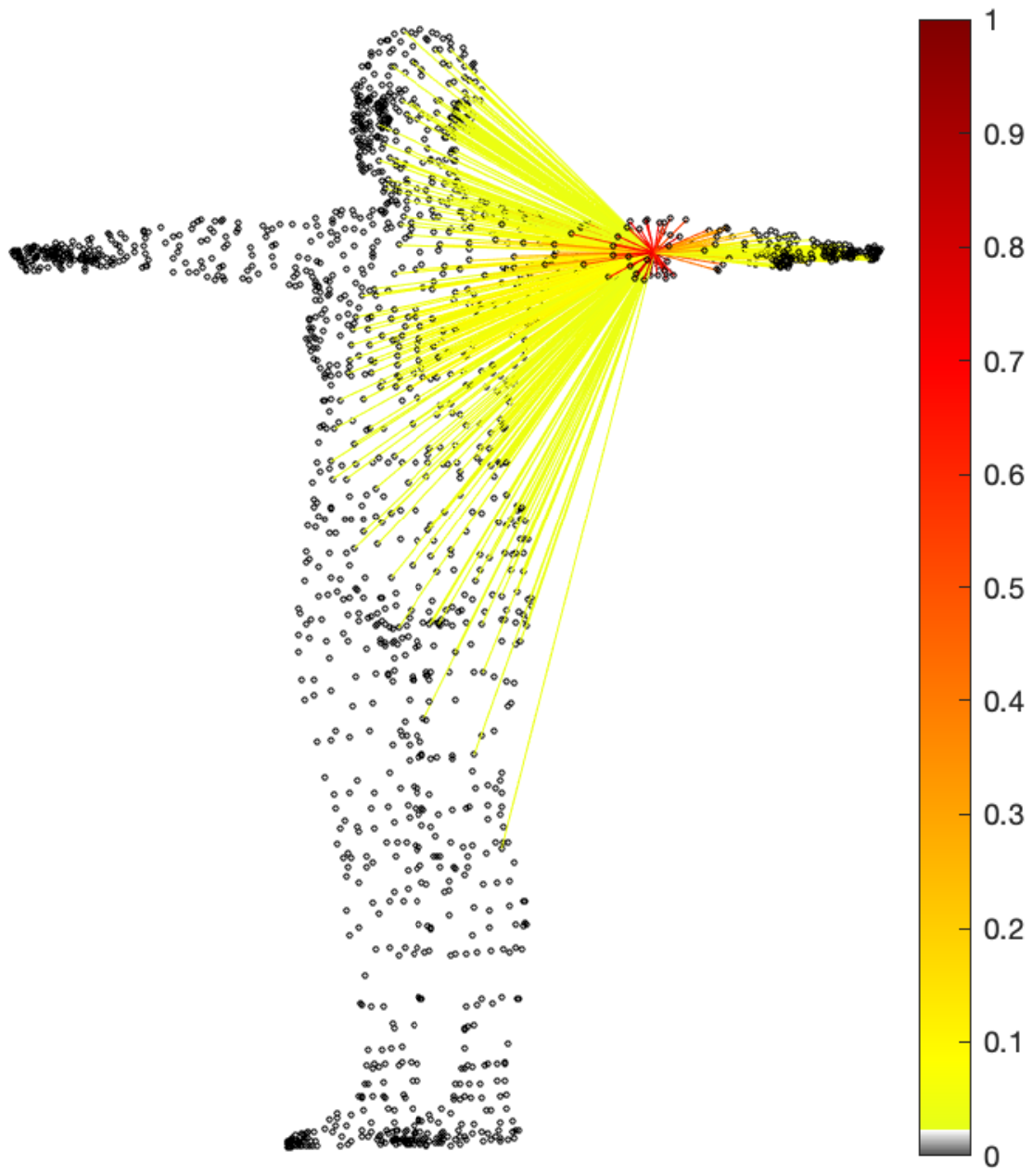}
        \includegraphics[width=0.99\linewidth]{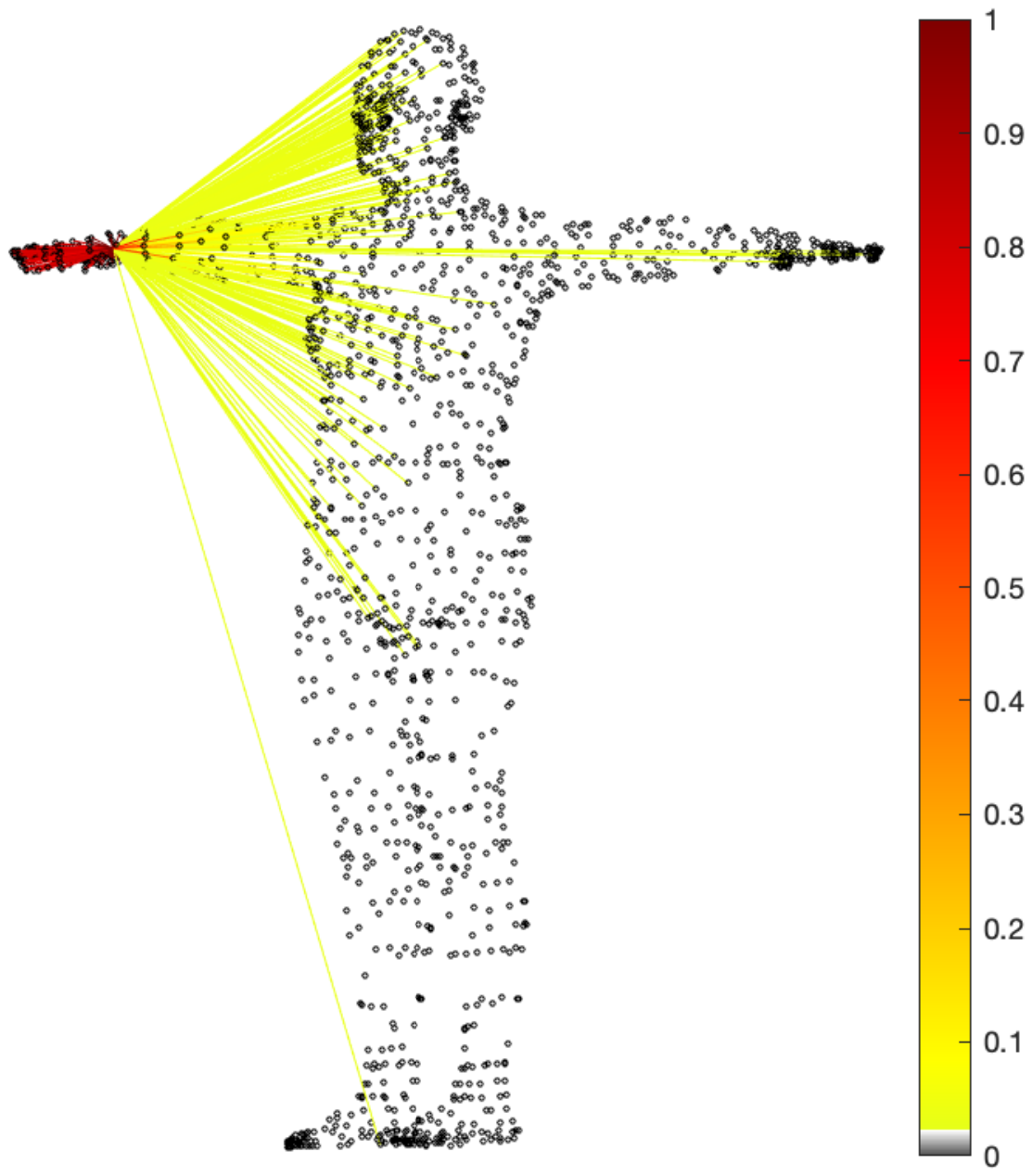}
    \end{subfigure}
    \begin{subfigure}{.12\textwidth}
         \includegraphics[width=0.99\linewidth]{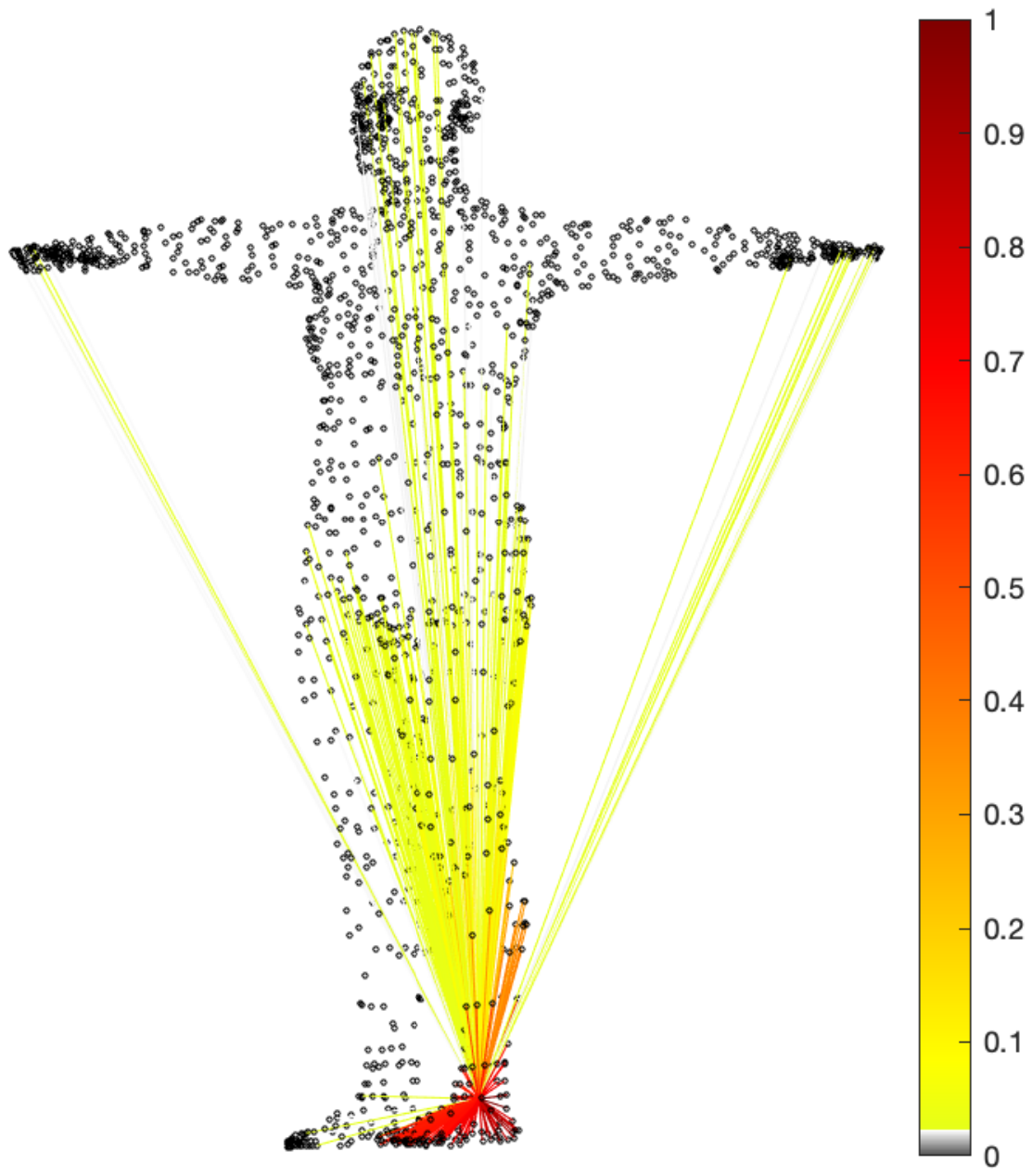}
        \includegraphics[width=0.99\linewidth]{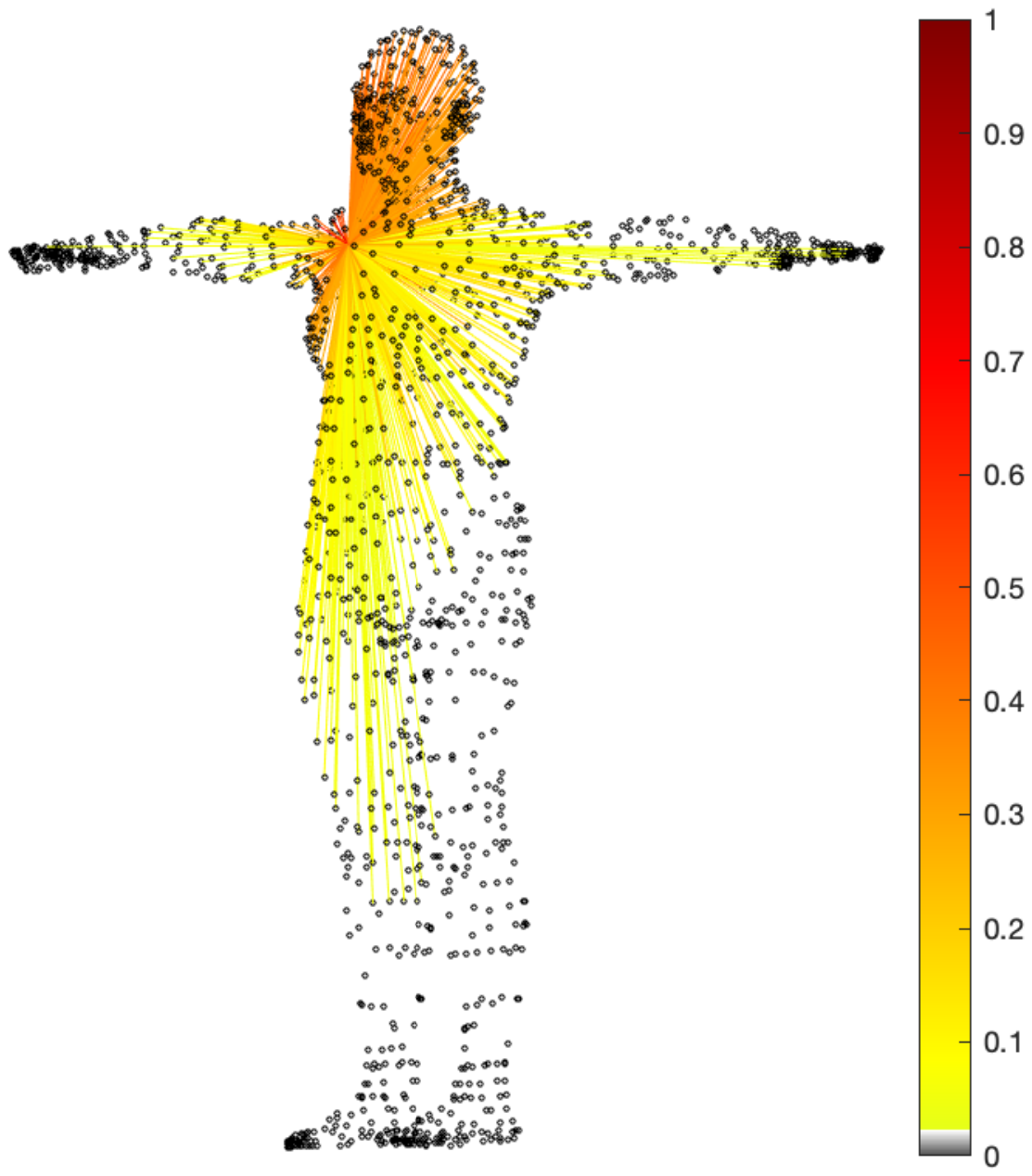}
    \end{subfigure}
\vspace{-2mm}
\caption{A visualization of the learned fusion adjacency matrix~($A_f$). \textbf{Left}: the connections between the right ankle and the mesh vertices. \textbf{Top row on the right}: right knee, left elbow, left ankle; \textbf{Bottom row on the right}: right hip, right wrist, right shoulder. Red, yellow and gray color indicate strong, weak and trivial connections.}
\label{fig:vis_fusion_A}
\end{figure*}

\section{Empirical Evaluations}
\label{sec:experiments}
We have evaluated our proposed method and present the results in this section. The datasets have different 2D annotations. We have selected the 12 joint annotations in common for the skeleton graph to define its graph structure. 

\subsection{Datasets and Evaluation metrics}
\noindent \textbf{Human 3.6M} This indoor 3D dataset~\cite{h36m_pami, IonescuSminchisescu11} comprises eleven subjects performing 17 common scenarios, e.g. sitting down, talking on the phone. The training data contains ground truth 2D joints, 3D joints, and SMPL (pose and shape) parameters. The entire dataset contains 3.6M images. For training we only have access to subjects 1, 5, 6, 7, and 8 (about 1.55M images). Subjects 9 and 11 are held out for evaluation (about 0.5M images). \\
\textbf{UP-3D} Unite the People 3D~\cite{Lassner:UP:2017} consists of images with annotations by humans doing sports and other miscellaneous activities. Besides ground truth 2d keypoints, SMPL fits have been performed on the 2D keypoints to produce ground truth SMPL parameters. About 7K images are used for training, and 639 held out for evaluation. \\
\textbf{LSP} Leeds Sports Pose~\cite{Johnson10} contains 2K images with 2D joint annotations of people playing sports. We use 1000 images for training, and 1000 for evaluation. \\
\textbf{COCO} Common Object in Context~\cite{MSCOCO_ECCV_2014} also contains images of people annotated 2D keypoints. About 28K images are used for training. We do not evaluate for this dataset. \\
\textbf{MPII} MPII Human Pose dataset contains images with annotated body joints of people performing 410 different activities~\cite{andriluka14cvpr}. We use about 15K training images from this dataset, and do not evaluate on this dataset. 

\noindent\textbf{Evaluation metrics}
For H36M we report the mean Euclidean distance~(\textbf{mm}) between the predicted and ground truth 3D joints after root joint alignment ({\it\bf MPJPE}), and rigid alignment error ({\it\bf PA-MPJPE}) as in ~\cite{Zhou:2019:MMHM}. For UP-3D we report {\it\bf MPVE}, which is a mean per-vertex error between the predicted and ground truth shape, and for LSP we report accuracy ({\it\bf Acc.}) and {\it\bf F1} score on foreground-background ({\it\bf FB seg}) and part segmentation ({\it\bf Parts seg}). We report non-parametric ({\it\bf np}) and SMPL parametric ({\it\bf p}) predictions for H36M P1, P2 and UP-3D datasets. 

\begin{table}[t]
\centering
\scriptsize
\begin{tabular}{l|c|c|c|c}
\hline
                                     & \multicolumn{2}{c|}{H36M P1} & \multicolumn{2}{c}{H36M P2}                 \\ \cline{2-3} \cline{4-5}
Methods                              & \tiny{MPJPE$\downarrow$}                    & \tiny{PA-MPJPE$\downarrow$}   & \tiny{MPJPE$\downarrow$}      & \tiny{PA-MPJPE$\downarrow$} \\ \hline
SMPLify~\cite{bogo2016keep}          & -                            & -          & -             & 82.3        \\
Lassner~\cite{Lassner:UP:2017}       & -                            & -          & -             & 93.9        \\
HMR~\cite{hmrKanazawa17}             & 88.5                         & 58.1       & -             & 56.8        \\
NBF~\cite{omran2018neural}           & -                            & -          & -             & 59.5        \\
Pavlakos~\cite{pavlakos2018learning} & -                            & -          & -             & 75.9        \\
Kanazawa~\cite{Kanazawa_2019_CVPR}   & -                            & -          & -             & 56.9        \\
Arnab~\cite{arnab2019exploiting}     & -                            & -          & 77.8          & 54.3        \\
GraphCMR~\cite{kolotouros2019cmr}    & 75.0                         & 51.2       & 72.7          & 49.3        \\
SPIN~\cite{Kolotouros_2019_ICCV}     & -                            & -          & -             & 41.1        \\
I2L-MeshNet~\cite{moon2020i2l}       & -                            & -          & 55.7          & 41.1        \\
METRO~\cite{lin2021end}              & -                            & -          & \textbf{54.0} & 36.7        \\
 \hline
Ours                                 & \textbf{61.2}                & \textbf{35.4} & 58.5       & \textbf{34.0} \\\hline
\end{tabular}
\caption{Comparison with the state-of-the-art on Human3.6M (Protocal 1 and 2) for estimated 3D poses (see suppl. mat. for per-activities results).}
\label{tbl:quan_eval_human3.6M}
\end{table}

\begin{table}[tb]
\centering
\scriptsize
\begin{tabular}{l|c|c|c|c}
\hline
\multirow{2}{*}{Methods} & \multicolumn{2}{c|}{FB seg} & \multicolumn{2}{c}{Parts seg} \\\cline{2-3} \cline{4-5}
                         & Acc.$\uparrow$                   & F1$\uparrow$ & Acc.$\uparrow$ & F1$\uparrow$         \\\hline
SMPLify oracle~\cite{bogo_eccv_2016}            & 92.17       &0.88 & 88.82 &0.67 \\
SMPLify~\cite{bogo_eccv_2016}                   & 91.89       &0.88 & 87.71 &0.64  \\
SMPLify on~\cite{pavlakos2018learning}                & 92.17       &0.88 & 88.24 &0.64 \\
Bodynet ~\cite{varol2018bodynet}                  & 92.75       &0.84  & - & -  \\
HMR~\cite{hmrKanazawa17}          & 91.67          & 0.87          & 87.12          & 0.60          \\
SPIN~\cite{Kolotouros_2019_ICCV}  & 91.83          & 0.87          & 89.41          & 0.68          \\
GraphCMR~\cite{kolotouros2019cmr} & 91.46          & 0.87          & 88.69          & 0.66          \\ \hline
Ours                              & \textbf{93.15} & \textbf{0.89} & \textbf{90.96} & \textbf{0.73} \\
\hline
\end{tabular}
\vspace*{-5pt}
\caption{Comparison with the state-of-the-art on LSP for 2D projection from the predicted non-parametric mesh. 
}
\label{tbl:quan_eval_lsp}
\end{table}

\begin{table}[t]
\centering
\scriptsize
\begin{tabular}{l|c|c}
\hline
Methods          & MPVE (np) $\downarrow$         & MPVE(p) $\downarrow$   \\ \hline
GraphCMR~\cite{kolotouros2019cmr} & 104.5        & 122.9        \\ 
Ours                              & \textbf{59.0} & \textbf{61.1} \\ \hline
\end{tabular}
\vspace*{-2mm}
\caption{Comparison with the mesh-only graph method~\cite{kolotouros2019cmr} on UP-3D for estimated 3D mes (MPVE is in mm). 
}
\label{tbl:quan_eval_up3d}
\end{table}

\subsection{Experiment Details}
\label{subsec:training}
We use a pre-trained ResNet-50 to extract perceptual features. Our model is trained end-to-end with a batch size of 64 and learning rate of \(2.5e^{-4}\). Mini-batches during training are assembled by selecting images from the five training datasets. The composition is 30\%, 20\%, 10\%, 20\% and 20\% for Human3.6M, UP-3D, LSP, COCO and MPII respectively. 
The Adam optimizer is used to determine the weight updates. We train our model for fifty epochs, but we observed fewer epochs could suffice (see Section~\ref{subsec:results}). 
During training, we apply the focal loss $\mathcal{L}_{fl}$ (with \(\alpha=1\), \(\gamma=1\)) to the estimated 3D pose from Mesh-GCN (\(\hat{\mathbf{V}}_s\)) and the coefficient before this loss term is 5.0. We mixed the ground truth and the estimated 2D joint location to get feature patches during training with a mixture ratio which gradually decreases to zero at the last epoch during training and only use the estimated 2D joints during inference. 



\subsection{Main Results and Analysis}
\label{subsec:results}
We compare our method to state-of-the-art methods on H36M, UP-3D and LSP datasets, which evaluates 3D poses, 3D mesh, and 2D projections of the mesh respectively. The results are shown in Tables~\ref{tbl:quan_eval_human3.6M},~\ref{tbl:quan_eval_up3d} and~\ref{tbl:quan_eval_lsp} for those three datasets. Our method either outperforms or achieves comparable performance as the prior methods on those datasets.

First of all, we aim to investigate how the bi-layer graph performs for body recovery. To this end, we first focus on the Human3.6M dataset and UP-3D dataset. The rich human activities in their images is a natural target to study the correlation between body parts, which requires long-range interactions. We evaluate the regressed mesh by our bi-layer graph through 3D pose accuracy, in comparison to the mesh-only graph method~\cite{Kolotouros_2019_CVPR} and the self-attention in transformer~\cite{lin2021end} as shown in Table~\ref{tbl:quan_eval_human3.6M}. In both cases, we outperform them in reconstruction error (PA-MPJPE), indicating that our proposed bi-layer graph uses the non-local interactions efficiently for body recovery. We also evaluate the regressed mesh and the mesh calculated from the regressed SMPL on the UP-3D dataset in  Table~\ref{tbl:quan_eval_up3d}, which demonstrates that our method can promote the fine-grained interactions between mesh vertices for improved body shape. We also evaluate 3D shape through silhouette projection on the LSP dataset in Table~\ref{tbl:quan_eval_lsp}. Our proposed bi-layered graph again outperforms prior methods. 

Our model aim to jointly model local vertex-vertex (defined by mesh neighbourhood), non-local vertex-joint, and joint-joint interactions. We get insight of vertex-joint intersections by the learned fusion adjacency matrix~($A_f$) in Fig.~\ref{fig:vis_fusion_A}. Firstly, strong interactions between a joint and its nearby mesh vertices are encouraged, thus the joint will guide the mesh recovery. We achieve this by setting the initial values of $A_f,s$~(see Eq.~\eqref{eq:Af}) as the reciprocal of the Euclidean vertex-join distance in T-pose mesh. 
This intersections cover a range larger than the fine-grained vertex-vertex interactions predefined by mesh neighborhood. 
Secondly, long-range joint-vertex interactions are learnt between a joint and remote vertices near another joint, when the body part correlation happens. Please see the example of the right ankle and right wrist in the left sub-figure of Fig. \ref{fig:vis_fusion_A}. 
Our intersections differ from the transformer-based METRO~\cite{lin2021end} in two ways: the local vertex-vertex intersections avoid huge computation of the brute-force self-attention; and joint-vertex intersections learnt from the Fusion Graph efficiently model the most important topology knowledge between body mesh and joints. Together with the localized image features, our model achieves comparable performance to METRO~\cite{lin2021end} of the strong representation ability for the fully connected intersections. We believe that attention and knowledge-aware bi-layer graph network can be integrated to learn the interactions.

\begin{figure}[!tbh]
\centering
    \begin{subfigure}[b]{0.47\textwidth}
        \includegraphics[width=\linewidth]{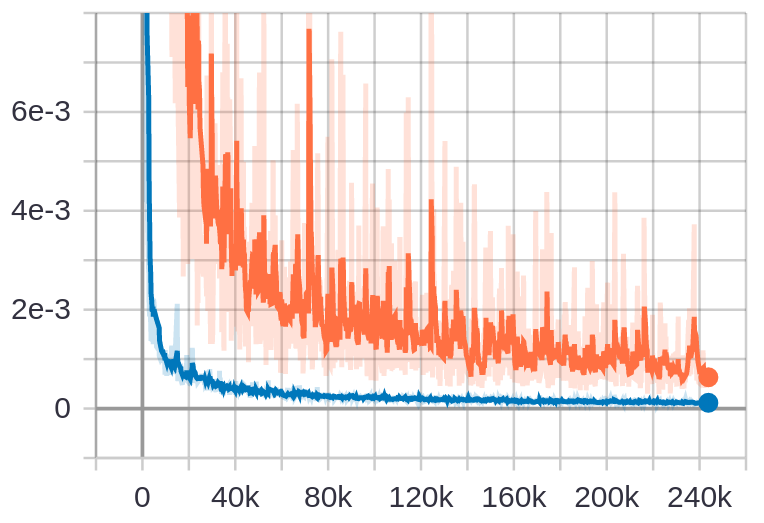}%
    \end{subfigure}
    \begin{subfigure}[b]{0.47\textwidth}
        \includegraphics[width=\linewidth]{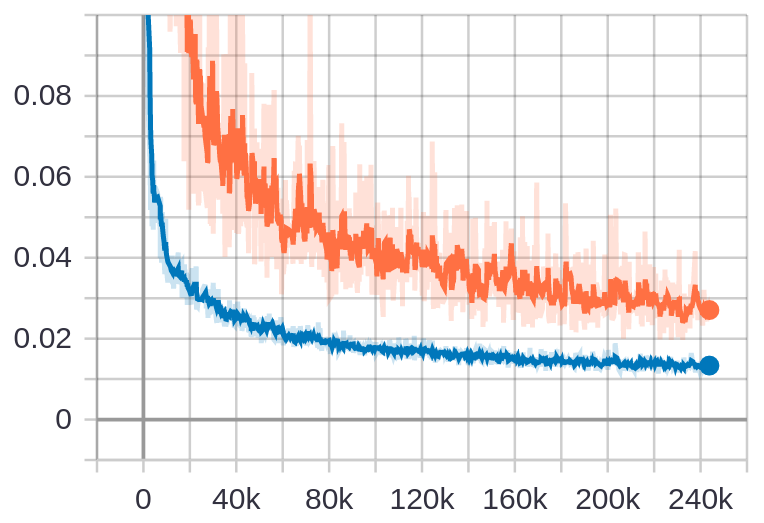}%
    \end{subfigure}    
\vspace{-2mm}
\caption{Comparison of  training loss on \(\hat{\mathcal{V}}^{j3d}_{m}\)(left) and \(\hat{\mathcal{V}}_{m}\)(right) from the common mesh graph structure of the \textcolor{tangelo}{ baseline}~\cite{kolotouros2019cmr}~ and our \textcolor{mediumtealblue}{proposed model}. We trained for 50 epochs and one epoch takes about 5,000 steps. 
}
\label{fig.visualize_convergence}
\end{figure}

Finally, we evaluate the efficiency of the bi-layer graph structure through the convergence speed of the training compared to the mesh-only method~\cite{kolotouros2019cmr} (see Fig.~\ref{fig.visualize_convergence}). Our model (in \textcolor{mediumtealblue}{blue}) achieves a lower, more stable loss much earlier compared to the baseline~\cite{kolotouros2019cmr} (in \textcolor{tangelo}{orange}), which indicates that the speedup of information propagation along this bi-layer graph can potentially reduce the training time.

\begin{figure}[!t]
    \centering
    \includegraphics[width=0.98\textwidth]{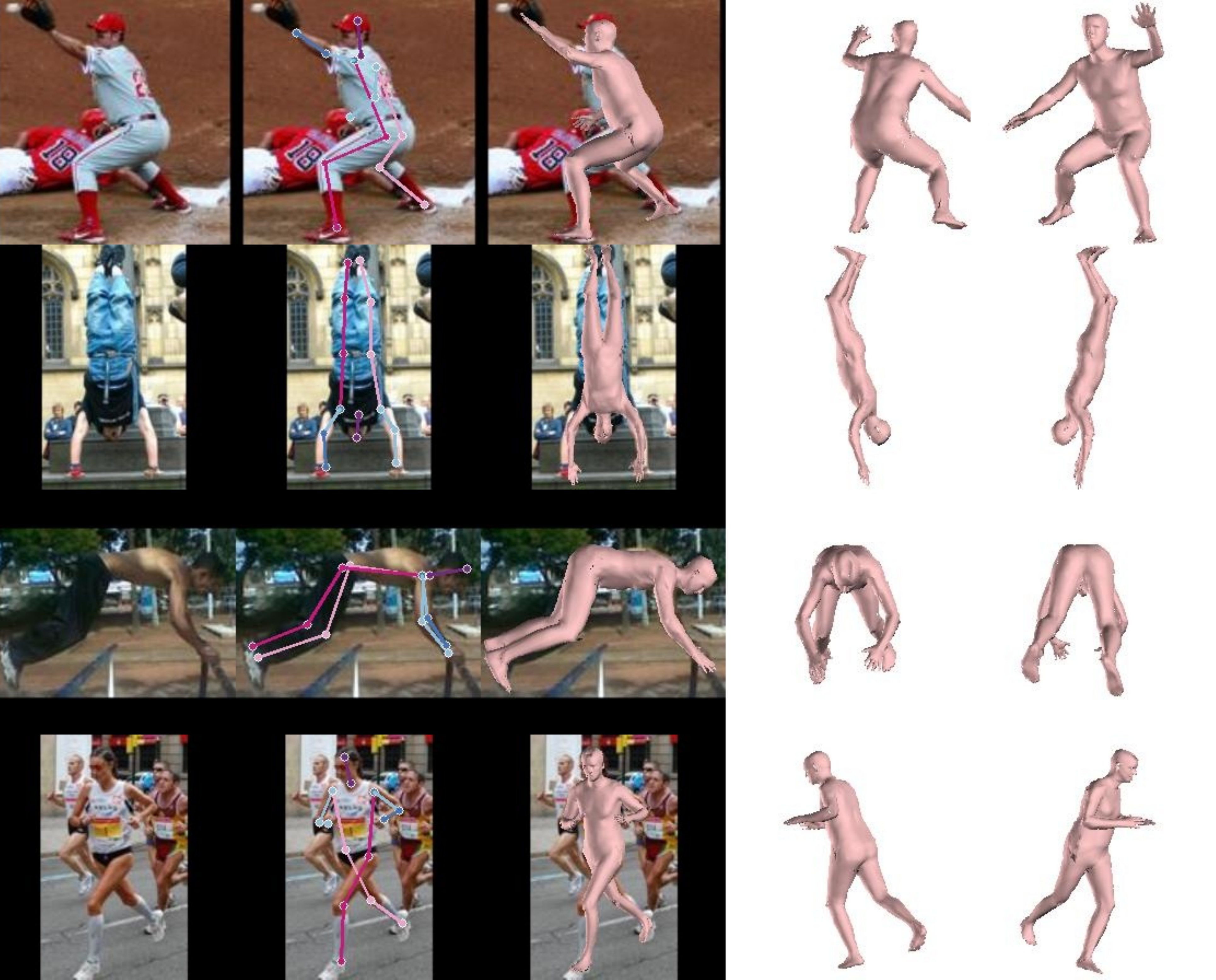}
    \caption{Qualitative results (left to right: input image, non-parametric pose and shape, two other view perspectives).}
    \label{fig:qualitative_results}
\end{figure}

\begin{figure}[!t]
    \centering
    \includegraphics[width=0.98\linewidth]{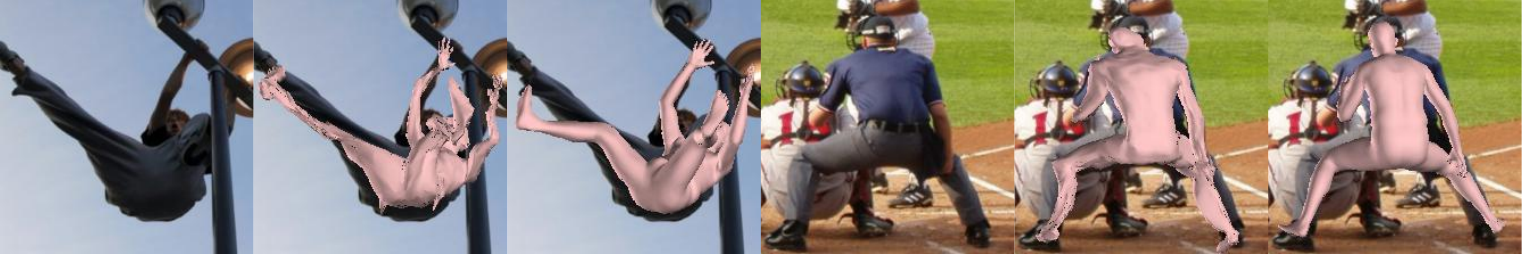}
    \caption{Two failure case examples (input, non-parametric, parametric).}
    \label{fig:failure_cases}
\end{figure}

\noindent\textbf{Qualitative Results} Fig.~\ref{fig:qualitative_results} and~\ref{fig:failure_cases} show four successful examples and  two failure cases due to challenging poses and occlusions (see suppl. mat. for more examples).



\subsection{Ablation Studies}
\label{subsec:ablation_studies}
\noindent\textbf{Benefit of bi-layer graph vs. localized features}
As shown in Table~\ref{tbl:ablation_arch}, instead of single-mesh only graph~(\textbf{A}), bi-layer graph only (\textbf{B}) or localized image features only (\textbf{C}) , it is the combination (\textbf{D})  of Fusion Graph  and localized features that jointly contribute most to the performance gain. 
This combination is the core difference from the transformer-based METRO~\cite{lin2021end} and other hierarchical structures, such as CoMA~\cite{ranjan2018generating} (see suppl. mat. for more discussion).

\noindent\textbf{Benefit of Fusion Graph} Since we design a bi-layer graph structure connected with fusion graph for mesh recovery, one interesting question is that whether the fusion graph is useful. In Table~\ref{tbl:ablation_results}, We study the fusion graph by limiting its usage in the bi-layer graph network: 
replacing with a simple fusion by pooling (\textbf{avgpool-as-fusion} and \textbf{maxpool-as-fusion}) and restricting it applied to only the first or the last graph layer(\textbf{fusion-at-first} and \textbf{fusion-at-last} respectively). 
The simple fusion strategy will loss the individual interaction between a joint vertex and a mesh vertex as each joint (mesh) vertex apply an identical feature from pooling the mesh (joint) vertices feature. We compare those strategies to our fusion graph and observe the performance increase significantly with allowing more fusion connections in the network and our fusion graph works on best with the full connections in all graph layers.

\noindent \textbf{Weight sharing} We exploit the property of GCN to share weights between skeleton-GCN and Mesh-GCN for compact model size. In Table~\ref{tbl:ablation_results}, it is interesting to observe only marginal performance loss, indicating the strong representation ability of GCN on the body and skeleton topology. 

\begin{table}[!t]
\centering
\scriptsize
\begin{tabular}{c|c|c|c|c|c}
\hline
 & Skeleton &  Fusion & Localized & H36M P2   & UP-3D \\  \cline{5-6}
&Graph & Graph & features & MPJPE $\downarrow$ & MPVE $\downarrow$      \\\hline
A&\xmark & \xmark & \xmark              & 54.0      & 104.5 \\
B&\cmark & \cmark & \xmark              & 47.5      & 96.3 \\
C&\cmark & \xmark & \cmark               & 48.8      &100.7\\
D&\cmark & \cmark & \cmark               & \textbf{34.0}     & \textbf{59.0}\\\hline
\end{tabular}
\vspace{-2mm}
\caption{\small{Evaluation of bi-layer graph components and localized features. All has Mesh Graph with global image features as input as GraphCMR and has the same training settings. }}
\label{tbl:ablation_arch}
\end{table}

\begin{table}[!t]
\centering
\scriptsize
\begin{tabular}{l|c|c}
\hline
 \multirow{2}{*}{Methods}                  & H36M P2   & UP-3D         \\  \cline{2-3}
                  & MPJPE $\downarrow$ & MPVE $\downarrow$      \\\hline
Ours              & \textbf{34.0}      & \textbf{59.0} \\\hline
avgpool-as-fusion & 47.3      & 81.1          \\
maxpool-as-fusion & 42.9      & 77.0          \\ 
fusion-at-first      & 36.7      & 64.3          \\
fusion-at-last       & 38.3      & 71.3          \\\hline
shared weight     & 34.2      & 61.7          \\
no FL             & 34.7      & 59.8          \\\hline
\end{tabular}
\vspace{-2mm}
\caption{Evaluation results for ablation studies. See text for details.
FL is for focal loss,
shared weight indicates the model shares weights between skeleton and mesh graph.
}
\label{tbl:ablation_results}
\end{table}

\noindent \textbf{Focal loss} To demonstrate the benefit of the focal loss for regression, we trained the model with $L_1$ loss on $\widehat{\textbf{V}}_s$ instead of the proposed focal loss and keep the other losses the same. In Table~\ref{tbl:ablation_results}, we see that $L_1$ loss works a bit inferior to the focal loss. We will explore the use of focal loss in future work for improving the overall performance. 

\section{Conclusion}
\label{sec:conclusion}

We have proposed a dual-scale graph-based method for 3D human shape and pose recovery from a single image. A skeleton graph estimates 3D pose, and a mesh graph estimates 3D shape. A fusion graph promotes the exchange of local and global information between the two graphs.
And Fusion Graph employs an adaptive adjacency matrix to learn which nodes between the two scales influence one another most. 
Our results show that we can outperform state-of-the-art methods. Some poses, and partial occlusions remain challenging. For future work, We would like to extend our work to take both single and multi-view images as input, which may help improve performance. 
In addition, 3D reconstruction of objects from images in general, not only humans, is an interesting research direction.

\paragraph{Acknowledgements}
This work was supported by MERL. Xin Yu was partially funded by the NSF grant IIS 1764071. 
We thank Srikumar Ramalingam for valuable discussions. 
We also thank the anonymous reviewers for their constructive feedback that helped in shaping the final manuscript.

{\small
\bibliographystyle{ieee_fullname}
\bibliography{cvprbib}

\begin{thebibliography}{10}\itemsep=-1pt

\bibitem{Alldieck_2019_ICCV}
Thiemo Alldieck, Gerard Pons-Moll, Christian Theobalt, and Marcus Magnor.
\newblock Tex2shape: Detailed full human body geometry from a single image.
\newblock In {\em Int. Conf. Comput. Vis.}, October 2019.

\bibitem{andriluka14cvpr}
Mykhaylo Andriluka, Leonid Pishchulin, Peter Gehler, and Bernt Schiele.
\newblock 2d human pose estimation: New benchmark and state of the art
  analysis.
\newblock In {\em IEEE Conf. Comput. Vis. Pattern Recog.}, June 2014.

\bibitem{arnab2019exploiting}
Anurag Arnab, Carl Doersch, and Andrew Zisserman.
\newblock Exploiting temporal context for 3d human pose estimation in the wild.
\newblock In {\em IEEE Conf. Comput. Vis. Pattern Recog.}, pages 3395--3404,
  2019.

\bibitem{bhatnagar2019multi}
Bharat~Lal Bhatnagar, Garvita Tiwari, Christian Theobalt, and Gerard Pons-Moll.
\newblock Multi-garment net: Learning to dress 3d people from images.
\newblock In {\em Int. Conf. Comput. Vis.}, pages 5420--5430, 2019.

\bibitem{bogo2016keep}
Federica Bogo, Angjoo Kanazawa, Christoph Lassner, Peter Gehler, Javier Romero,
  and Michael~J Black.
\newblock Keep it smpl: Automatic estimation of 3d human pose and shape from a
  single image.
\newblock In {\em Eur. Conf. Comput. Vis.}, pages 561--578. Springer, 2016.

\bibitem{bogo_eccv_2016}
Federica Bogo, Angjoo Kanazawa, Christoph Lassner, Peter Gehler, Javier Romero,
  and Michael~J. Black.
\newblock Keep it smpl: Automatic estimation of 3d human pose and shape from a
  single image.
\newblock In Bastian Leibe, Jiri Matas, Nicu Sebe, and Max Welling, editors,
  {\em Eur. Conf. Comput. Vis.}, pages 561--578, Cham, 2016. Springer
  International Publishing.

\bibitem{cao2018openpose}
Zhe Cao, Gines Hidalgo, Tomas Simon, Shih-En Wei, and Yaser Sheikh.
\newblock Open{P}ose: realtime multi-person 2{D} pose estimation using {P}art
  {A}ffinity {F}ields.
\newblock In {\em arXiv preprint arXiv:1812.08008}, 2018.

\bibitem{Choi_2020}
Hongsuk Choi, Gyeongsik Moon, and Kyoung~Mu Lee.
\newblock Pose2mesh: Graph convolutional network for 3d human pose and mesh
  recovery from a 2d human pose.
\newblock {\em IEEE Conf. Comput. Vis. Pattern Recog.}, 2020.

\bibitem{Ci_2019_ICCV}
Hai Ci, Chunyu Wang, Xiaoxuan Ma, and Yizhou Wang.
\newblock Optimizing network structure for 3d human pose estimation.
\newblock In {\em Int. Conf. Comput. Vis.}, October 2019.

\bibitem{Gabeur_2019_ICCV}
Valentin Gabeur, Jean-Sebastien Franco, Xavier Martin, Cordelia Schmid, and
  Gregory Rogez.
\newblock Moulding humans: Non-parametric 3d human shape estimation from single
  images.
\newblock In {\em Int. Conf. Comput. Vis.}, October 2019.

\bibitem{meshrcnn}
Justin~Johnson Georgia~Gkioxari, Jitendra~Malik.
\newblock Mesh r-cnn.
\newblock {\em Int. Conf. Comput. Vis.}, 2019.

\bibitem{Gong_2019_CVPR}
Ke Gong, Yiming Gao, Xiaodan Liang, Xiaohui Shen, Meng Wang, and Liang Lin.
\newblock Graphonomy: Universal human parsing via graph transfer learning.
\newblock In {\em IEEE Conf. Comput. Vis. Pattern Recog.}, June 2019.

\bibitem{he_mask_rcnn}
K. {He}, G. {Gkioxari}, P. {Dollár}, and R. {Girshick}.
\newblock Mask r-cnn.
\newblock In {\em Int. Conf. Comput. Vis.}, pages 2980--2988, Oct 2017.

\bibitem{he2016deep}
Kaiming He, Xiangyu Zhang, Shaoqing Ren, and Jian Sun.
\newblock Deep residual learning for image recognition.
\newblock In {\em IEEE Conf. Comput. Vis. Pattern Recog.}, pages 770--778,
  2016.

\bibitem{IonescuSminchisescu11}
Catalin Ionescu, Fuxin Li, and Cristian Sminchisescu.
\newblock Latent structured models for human pose estimation.
\newblock In {\em 2011 International Conference on Computer Vision}, pages
  2220--2227. IEEE, 2011.

\bibitem{h36m_pami}
Catalin Ionescu, Dragos Papava, Vlad Olaru, and Cristian Sminchisescu.
\newblock Human3.6m: Large scale datasets and predictive methods for 3d human
  sensing in natural environments.
\newblock {\em IEEE Trans. Pattern Anal. Mach. Intell.}, 36(7):1325--1339, jul
  2014.

\bibitem{jiang2020bcnet}
Boyi Jiang, Juyong Zhang, Yang Hong, Jinhao Luo, Ligang Liu, and Hujun Bao.
\newblock Bcnet: Learning body and cloth shape from a single image.
\newblock {\em Eur. Conf. Comput. Vis.}, 2020.

\bibitem{Johnson10}
Sam Johnson and Mark Everingham.
\newblock Clustered pose and nonlinear appearance models for human pose
  estimation.
\newblock In {\em Brit. Mach. Vis. Conf.}, 2010.
\newblock doi:10.5244/C.24.12.

\bibitem{hmrKanazawa17}
Angjoo Kanazawa, Michael~J. Black, David~W. Jacobs, and Jitendra Malik.
\newblock End-to-end recovery of human shape and pose.
\newblock In {\em IEEE Conf. Comput. Vis. Pattern Recog.}, 2018.

\bibitem{Kanazawa_2018_CVPR}
Angjoo Kanazawa, Michael~J. Black, David~W. Jacobs, and Jitendra Malik.
\newblock End-to-end recovery of human shape and pose.
\newblock In {\em Proceedings of the IEEE Conference on Computer Vision and
  Pattern Recognition (CVPR)}, June 2018.

\bibitem{Kanazawa_2019_CVPR}
Angjoo Kanazawa, Jason~Y. Zhang, Panna Felsen, and Jitendra Malik.
\newblock Learning 3d human dynamics from video.
\newblock In {\em IEEE Conf. Comput. Vis. Pattern Recog.}, June 2019.

\bibitem{kavan2014part}
Ladislav Kavan.
\newblock Part i: direct skinning methods and deformation primitives.
\newblock In {\em ACM SIGGRAPH}, volume 2014, pages 1--11, 2014.

\bibitem{Kipf:2016tc}
Thomas~N. Kipf and Max Welling.
\newblock {Semi-Supervised Classification with Graph Convolutional Networks}.
\newblock In {\em Proceedings of the 5th International Conference on Learning
  Representations}, ICLR '17, 2017.

\bibitem{Kolotouros_2019_ICCV}
Nikos Kolotouros, Georgios Pavlakos, Michael~J. Black, and Kostas Daniilidis.
\newblock Learning to reconstruct 3d human pose and shape via model-fitting in
  the loop.
\newblock In {\em Int. Conf. Comput. Vis.}, October 2019.

\bibitem{kolotouros2019cmr}
Nikos Kolotouros, Georgios Pavlakos, and Kostas Daniilidis.
\newblock Convolutional mesh regression for single-image human shape
  reconstruction.
\newblock In {\em IEEE Conf. Comput. Vis. Pattern Recog.}, 2019.

\bibitem{Kolotouros_2019_CVPR}
Nikos Kolotouros, Georgios Pavlakos, and Kostas Daniilidis.
\newblock Convolutional mesh regression for single-image human shape
  reconstruction.
\newblock In {\em Proceedings of the IEEE/CVF Conference on Computer Vision and
  Pattern Recognition (CVPR)}, June 2019.

\bibitem{Lassner:UP:2017}
Christoph Lassner, Javier Romero, Martin Kiefel, Federica Bogo, Michael~J.
  Black, and Peter~V. Gehler.
\newblock Unite the people: Closing the loop between 3d and 2d human
  representations.
\newblock In {\em IEEE Conf. Comput. Vis. Pattern Recog.}, July 2017.

\bibitem{li2019symbiotic}
Maosen Li, Siheng Chen, Xu Chen, Ya Zhang, Yanfeng Wang, and Qi Tian.
\newblock Symbiotic graph neural networks for 3d skeleton-based human action
  recognition and motion prediction.
\newblock arXiv, 2019.

\bibitem{lin2021end}
Kevin Lin, Lijuan Wang, and Zicheng Liu.
\newblock End-to-end human pose and mesh reconstruction with transformers.
\newblock In {\em Proceedings of the IEEE/CVF Conference on Computer Vision and
  Pattern Recognition}, pages 1954--1963, 2021.

\bibitem{2017FocalLF}
Tsung{-}Yi Lin, Priya Goyal, Ross~B. Girshick, Kaiming He, and Piotr
  Doll{\'{a}}r.
\newblock Focal loss for dense object detection.
\newblock {\em Int. Conf. Comput. Vis.}, pages 2999--3007, 2017.

\bibitem{MSCOCO_ECCV_2014}
Tsung-Yi Lin, Michael Maire, Serge Belongie, James Hays, Pietro Perona, Deva
  Ramanan, Piotr Doll{\'a}r, and C~Lawrence Zitnick.
\newblock Microsoft coco: Common objects in context.
\newblock In {\em European conference on computer vision}, pages 740--755.
  Springer, 2014.

\bibitem{SMPL:2015}
Matthew Loper, Naureen Mahmood, Javier Romero, Gerard Pons-Moll, and Michael~J.
  Black.
\newblock {SMPL}: A skinned multi-person linear model.
\newblock {\em ACM Trans. Graphics (Proc. SIGGRAPH Asia)}, 34(6):248:1--248:16,
  Oct. 2015.

\bibitem{moon2020i2l}
Gyeongsik Moon and Kyoung~Mu Lee.
\newblock I2l-meshnet: Image-to-lixel prediction network for accurate 3d human
  pose and mesh estimation from a single rgb image.
\newblock {\em Eur. Conf. Comput. Vis.}, 2020.

\bibitem{omran2018neural}
Mohamed Omran, Christoph Lassner, Gerard Pons-Moll, Peter Gehler, and Bernt
  Schiele.
\newblock Neural body fitting: Unifying deep learning and model based human
  pose and shape estimation.
\newblock In {\em IEEE Int. Conf. on 3D Vision (3DV)}, pages 484--494, 2018.

\bibitem{Pandey_2019_CVPR}
Rohit Pandey, Anastasia Tkach, Shuoran Yang, Pavel Pidlypenskyi, Jonathan
  Taylor, Ricardo Martin-Brualla, Andrea Tagliasacchi, George Papandreou,
  Philip Davidson, Cem Keskin, Shahram Izadi, and Sean Fanello.
\newblock Volumetric capture of humans with a single rgbd camera via
  semi-parametric learning.
\newblock In {\em IEEE Conf. Comput. Vis. Pattern Recog.}, June 2019.

\bibitem{Pavlakos_2019_ICCV}
Georgios Pavlakos, Nikos Kolotouros, and Kostas Daniilidis.
\newblock Texturepose: Supervising human mesh estimation with texture
  consistency.
\newblock In {\em Int. Conf. Comput. Vis.}, October 2019.

\bibitem{pavlakos2018learning}
Georgios Pavlakos, Luyang Zhu, Xiaowei Zhou, and Kostas Daniilidis.
\newblock Learning to estimate 3d human pose and shape from a single color
  image.
\newblock In {\em IEEE Conf. Comput. Vis. Pattern Recog.}, pages 459--468,
  2018.

\bibitem{Pumarola_2019_ICCV}
Albert Pumarola, Jordi Sanchez-Riera, Gary P.~T. Choi, Alberto Sanfeliu, and
  Francesc Moreno-Noguer.
\newblock 3dpeople: Modeling the geometry of dressed humans.
\newblock In {\em Int. Conf. Comput. Vis.}, October 2019.

\bibitem{ranjan2018generating}
Anurag Ranjan, Timo Bolkart, Soubhik Sanyal, and Michael~J Black.
\newblock Generating 3d faces using convolutional mesh autoencoders.
\newblock In {\em Proceedings of the European Conference on Computer Vision
  (ECCV)}, pages 704--720, 2018.

\bibitem{pifuSHNMKL19}
Shunsuke Saito, , Zeng Huang, Ryota Natsume, Shigeo Morishima, Angjoo Kanazawa,
  and Hao Li.
\newblock Pifu: Pixel-aligned implicit function for high-resolution clothed
  human digitization.
\newblock {\em Int. Conf. Comput. Vis.}, 2019.

\bibitem{Saito2020:PifuHD}
Shunsuke Saito, Tomas Simon, Jason Saragih, and Hanbyul Joo.
\newblock Pifuhd: Multi-level pixel-aligned implicit function for
  high-resolution 3d human digitization.
\newblock {\em IEEE Conf. Comput. Vis. Pattern Recog.}, 2020.

\bibitem{sun2019deep}
Ke Sun, Bin Xiao, Dong Liu, and Jingdong Wang.
\newblock Deep high-resolution representation learning for human pose
  estimation.
\newblock In {\em IEEE Conf. Comput. Vis. Pattern Recog.}, pages 5693--5703,
  2019.

\bibitem{Sun_2019_ICCV}
Yu Sun, Yun Ye, Wu Liu, Wenpeng Gao, Yili Fu, and Tao Mei.
\newblock Human mesh recovery from monocular images via a skeleton-disentangled
  representation.
\newblock In {\em Int. Conf. Comput. Vis.}, October 2019.

\bibitem{varol2018bodynet}
Gul Varol, Duygu Ceylan, Bryan Russell, Jimei Yang, Ersin Yumer, Ivan Laptev,
  and Cordelia Schmid.
\newblock Bodynet: Volumetric inference of 3d human body shapes.
\newblock In {\em Proceedings of the European Conference on Computer Vision
  (ECCV)}, pages 20--36, 2018.

\bibitem{Wandt_2019_CVPR}
Bastian Wandt and Bodo Rosenhahn.
\newblock Repnet: Weakly supervised training of an adversarial reprojection
  network for 3d human pose estimation.
\newblock In {\em IEEE Conf. Comput. Vis. Pattern Recog.}, June 2019.

\bibitem{Wang_2018_ECCV}
Nanyang Wang, Yinda Zhang, Zhuwen Li, Yanwei Fu, Wei Liu, and Yu-Gang Jiang.
\newblock Pixel2mesh: Generating 3d mesh models from single rgb images.
\newblock {\em Eur. Conf. Comput. Vis.}, 04 2018.

\bibitem{Wen_2019_ICCV}
Chao Wen, Yinda Zhang, Zhuwen Li, and Yanwei Fu.
\newblock Pixel2mesh++: Multi-view 3d mesh generation via deformation.
\newblock In {\em Int. Conf. Comput. Vis.}, October 2019.

\bibitem{Wu_2018_ECCV}
Yuxin Wu and Kaiming He.
\newblock Group normalization.
\newblock In {\em Eur. Conf. Comput. Vis.}, September 2018.

\bibitem{Yu_2019_CVPR}
Tao Yu, Zerong Zheng, Yuan Zhong, Jianhui Zhao, Qionghai Dai, Gerard Pons-Moll,
  and Yebin Liu.
\newblock Simulcap : Single-view human performance capture with cloth
  simulation.
\newblock In {\em IEEE Conf. Comput. Vis. Pattern Recog.}, June 2019.

\bibitem{Zhang_2019_CVPR}
Feng Zhang, Xiatian Zhu, and Mao Ye.
\newblock Fast human pose estimation.
\newblock In {\em IEEE Conf. Comput. Vis. Pattern Recog.}, June 2019.

\bibitem{Zhang_2019_ICCV}
Jason~Y. Zhang, Panna Felsen, Angjoo Kanazawa, and Jitendra Malik.
\newblock Predicting 3d human dynamics from video.
\newblock In {\em Int. Conf. Comput. Vis.}, October 2019.

\bibitem{Zhao_2019_CVPR}
Long Zhao, Xi Peng, Yu Tian, Mubbasir Kapadia, and Dimitris~N. Metaxas.
\newblock Semantic graph convolutional networks for 3d human pose regression.
\newblock In {\em IEEE Conf. Comput. Vis. Pattern Recog.}, June 2019.

\bibitem{Zheng_2019_ICCV}
Zerong Zheng, Tao Yu, Yixuan Wei, Qionghai Dai, and Yebin Liu.
\newblock Deephuman: 3d human reconstruction from a single image.
\newblock In {\em Int. Conf. Comput. Vis.}, October 2019.

\bibitem{Zhou:2019:MMHM}
Xiaowei Zhou, Menglong Zhu, Georgios Pavlakos, Spyridon Leonardos,
  Konstantinos~G. Derpanis, and Kostas Daniilidis.
\newblock Monocap: Monocular human motion capture using a cnn coupled with a
  geometric prior.
\newblock {\em IEEE Trans. Pattern Anal. Mach. Intell.}, 41(4):901--914, Apr.
  2019.

\end{thebibliography}
}

\newpage
\clearpage

\appendix

\section*{Additional Studies and Results}
\label{sec:supplementary}

\subsection*{Per-Activity Evaluation for Human 3.6M}

The Human 3.6M dataset~\cite{h36m_pami, IonescuSminchisescu11} contains people performing 15 activities, such as sitting down and walking. In the main paper, we reported our best model (shown in Table 2 of the main paper), which applies No Weight Sharing
. Table~\ref{tbl:per_activity_results_1} and~\ref{tbl:per_activity_results_2} shows the evaluation of non-parametric and SMPL parametric predictions on Human 3.6M for each activity separately. The activities represent Providing Directions, Having a Discussion, Eating, Greeting, Making a Phone Call, Taking a Photo, Posing, Making a Purchase, Sitting, Sitting Down, Smoking, Waiting, Walking a Dog, Walking Together, and Walking respectively. It is clear from the table that the performance for certain activities (highlighted in red), e.g., Walking, compares favorably to others, e.g., Sitting Down, as those poses are much more challenging and consequently have higher errors.

\begin{table}[h]
\centering
\scriptsize
\begin{tabular}{l | c | c | c | c}
  \hline
   \multicolumn{1}{l|}{}          & \multicolumn{2}{c|}{P1} & \multicolumn{2}{c}{P2}  \\ \hline
     Act.                          & \scriptsize{MPJPE}\tiny{$\downarrow$} & \scriptsize{PA-MPJPE}\tiny{$\downarrow$} & \scriptsize{MPJPE}\tiny{$\downarrow$} & \scriptsize{PA-MPJPE}\tiny{$\downarrow$} \\ \hline
  \textcolor{red}{Directions}                      & 61.29          & 31.61          & 55.24          & \textbf{28.14} \\
  \textcolor{red}{Discussion}                      & 59.62          & 33.84          & 56.19          & 32.30          \\
  Eating                          & 58.43          & 34.46          & 55.34          & 33.51          \\
  Greeting                        & 62.68          & 34.81          & 58.36          & 33.53          \\
  Phoning                         & 60.32          & 35.48          & 58.14          & 33.50          \\
  Photo         & 68.23          & 39.27          & 63.66          & 38.23          \\
  \textcolor{red}{Posing}       & 61.22          & 33.43          & 57.86          & 29.98          \\
  \textcolor{red}{Purchases}                       & 60.59          & 32.18          & 59.39          & 31.75          \\
    Sitting        & 66.63          & 40.51          & 64.92          & 42.21          \\
  SittingDown   & 72.74          & 49.57          & 72.18          & 45.44          \\
   Smoking      & 57.77          & 34.09          & 54.06          & 33.21          \\
   \textcolor{red}{Waiting}      & 61.80          & 34.23          & 57.93          & 31.51          \\
   WalkDog                        & 58.09          & 34.50          & 59.16          & 35.31          \\
   \textcolor{red}{WalkTogether} & 57.36          & 31.41          & 56.49          & 30.79          \\
   \textcolor{red}{Walking}      & \textbf{52.72} & \textbf{29.46} & \textbf{52.03} & 28.66          \\
   Overall                        & 61.17          & 35.36          & 58.45          & 33.96          \\
  \hline
\end{tabular}
\vspace*{5pt}
\caption{Evaluation of non-parametric predictions on Human 3.6M per activity. Certain activities (in red) result in better performance, compared to others. Numbers are MPJPE and PA-MPJPE in mm.}
\label{tbl:per_activity_results_1}
\end{table}

\begin{table}[h]
\centering
\scriptsize
\begin{tabular}{l | c | c | c | c}
  \hline
   \multicolumn{1}{l|}{}          & \multicolumn{2}{c|}{P1} & \multicolumn{2}{c}{P2}  \\ \hline
     Act.                          & \scriptsize{MPJPE}\tiny{$\downarrow$} & \scriptsize{PA-MPJPE}\tiny{$\downarrow$} & \scriptsize{MPJPE}\tiny{$\downarrow$} & \scriptsize{PA-MPJPE}\tiny{$\downarrow$} \\ \hline
  \textcolor{red}{Directions}                      & 62.65                   & 34.72                      & 56.83                   & 32.11                      \\
  Discussion                      & 62.00                   & 37.43                      & 58.27                   & 36.44                      \\
  Eating                          & 63.35                   & 38.59                      & 59.95                   & 37.32                      \\
  Greeting                        & 64.49                   & 38.63                      & 59.96                   & 37.30                      \\
  Phoning                         & 65.71                   & 40.39                      & 63.45                   & 37.92                      \\
  Photo          & 74.37                   & 45.42                      & 69.97                   & 44.86                      \\
  Posing                          & 64.06                   & 37.97                      & 59.08                   & 34.96                      \\
  Purchases                       & 65.84                   & 37.47                      & 62.52                   & 37.46                      \\
  Sitting       & 74.62                   & 47.14                      & 70.46                   & 46.33                      \\
   SittingDown   & 79.76                   & 53.81                      & 79.31                   & 51.85                      \\
   Smoking      & 62.57                   & 40.04                      & 59.10                   & 38.79                      \\
   Waiting                        & 64.18                   & 37.97                      & 59.42                   & 35.74                      \\
   WalkDog                        & 63.57                   & 39.57                      & 63.71                   & 40.44                      \\
   \textcolor{red}{WalkTogether} & 59.60                   & 34.73                      & 58.83                   & 34.15                      \\
   \textcolor{red}{Walking}      & \textbf{55.09}          & \textbf{32.87}             & \textbf{54.02}          & \textbf{32.05}             \\
   Overall                        & 65.35                   & 39.91                      & 62.16                   & 38.56                      \\
  \hline
\end{tabular}
\vspace*{5pt}
\caption{Evaluation of SMPL parametric prediction on Human 3.6M per activity. Certain activities (in red) result in better performance, compared to others.  Numbers are MPJPE and PA-MPJPE in mm.}
\label{tbl:per_activity_results_2}
\end{table}

\subsection*{Non-Parametric vs Parametric}
\noindent\textbf{Result comparison} As shown in Table \ref{tbl:per_activity_results_1} and \ref{tbl:per_activity_results_2}, the non-parametric shape is the regression result of all mesh vertices from the Bi-layer Graph and is generally able to learn the pose better than the SMPL parametric predictions on Human 3.6M dataset. Please note the evaluation on SMPL parametric predictions is still comparable to the state-of-the-arts as shown in Table 1 in the main paper. As introduced in the main paper, the Bi-layer Graph and SMPL regressor forms a pipeline, and the input of the later module depends on the output of the former module. The good performance of the later module further illustrates that our proposed Bi-layer Graph has addressed the dense regression of body mesh vertices well. Comparing the third and last rows of Figure~\ref{fig:qualitative_results} and Figure~\ref{fig:qualitative_results_smpl}, we can observe that some keypoints of the SMPL prediction, e.g. ankles, are slightly off the ground truth while the non-parametric poses are more accurate.

\subsection*{More discussion about other networks}
\noindent\textbf{METRO} Both the transformer-based METRO~\cite{lin2021end} and our model aim to jointly model vertex-vertex, vertex-joint, and joint-joint interactions. But they differ in the ways of representing each vertex and joint and learning those interactions. 
METRO uses self-attention to brute-force learn all interactions.
Although powerful, 
the self-attention has a well-known issue of the quadratic time and memory complexity. METRO has to down-sample the mesh to 431 vertices and train on very large mixed datasets for a long time~(200 epochs) to learn all the interactions. 
Rather than the brute-force self-attention, we 
inject prior knowledge of the mesh topology into the Mesh Graph~(1723 vertices), whose adjacency matrix is naturally sparse. In this way, our model  trains on a smaller amount of data for 50 epochs and converges faster: the accuracy increases rapidly in the first 12 epochs and becomes stable after 32 epochs as shown in Figure 5 in the main paper. Together with the localized image features, our model achieves comparable performance to self-attention based METRO. We believe that attention and knowledge-aware bi-layer graph network can be integrated to learn the interactions.

\noindent\textbf{CoMA}
Compared to the hierarchy GCN capturing face shape and expression at multiple scales in CoMA~\cite{ranjan2018generating}, our bi-layer graph is simple and efficient to represent non-local body mesh with the additional skeleton-scale graph. Firstly, we use the prior knowledge that the body mesh highly depends on the joint motion. Secondly, extra intermediate-scale body mesh representations by down-sampling (as in CoMA) doesn't help based on our trials. Additionally, our bi-layer graph uses both vertex and joint inputs, rather than just vertices as CoMA does. Our Fusion Graph further learns dynamic vertex-joint correlations, while the transform matrices of down-sampling and up-sampling layers in CoMA are predefined and fixed. %


\begin{figure}[h!]
    \centering
    \includegraphics[width=0.98\textwidth]{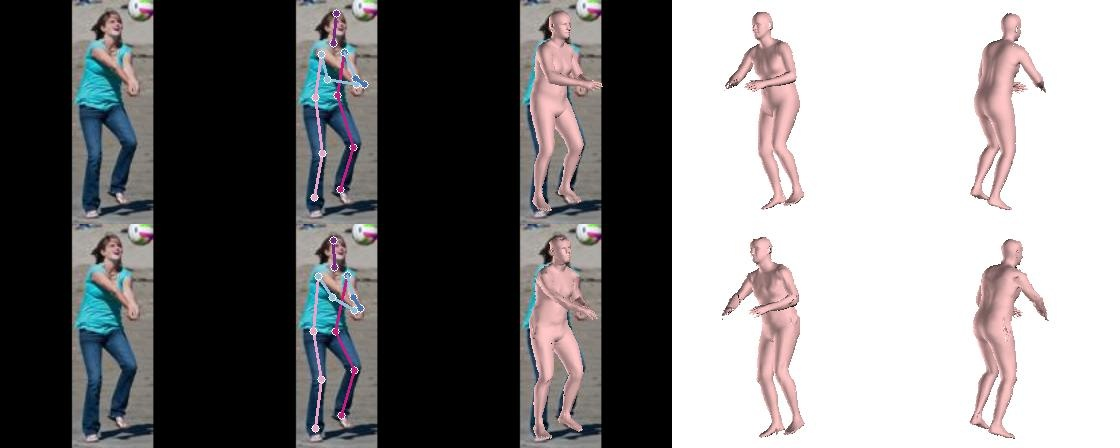}
    \caption{One example of bad pose in different 3D views. The two rows show parametric and non-parametric results respectively.}
    \label{fig:extra_qualitative_results_bad}
\end{figure}

\begin{figure*}[t]
    \centering
    \includegraphics[width=0.7\textwidth]{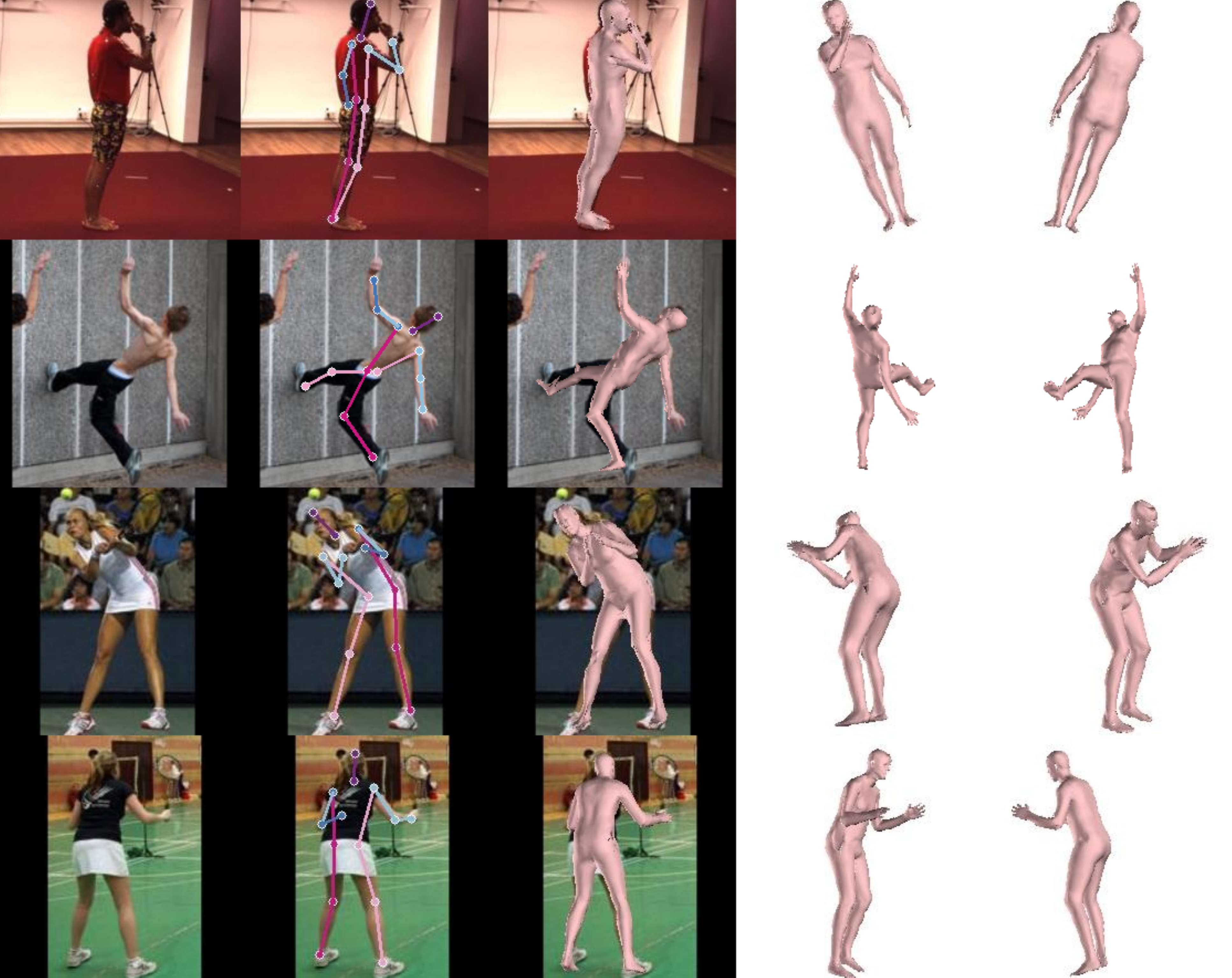}
    \caption{Qualitative non-parametric results. From left to right: input image, pose and shape, two different view perspectives.}
    \label{fig:extra_qualitative_results}
\end{figure*}

\begin{figure*}[h]
    \centering
    \includegraphics[width=0.7\textwidth]{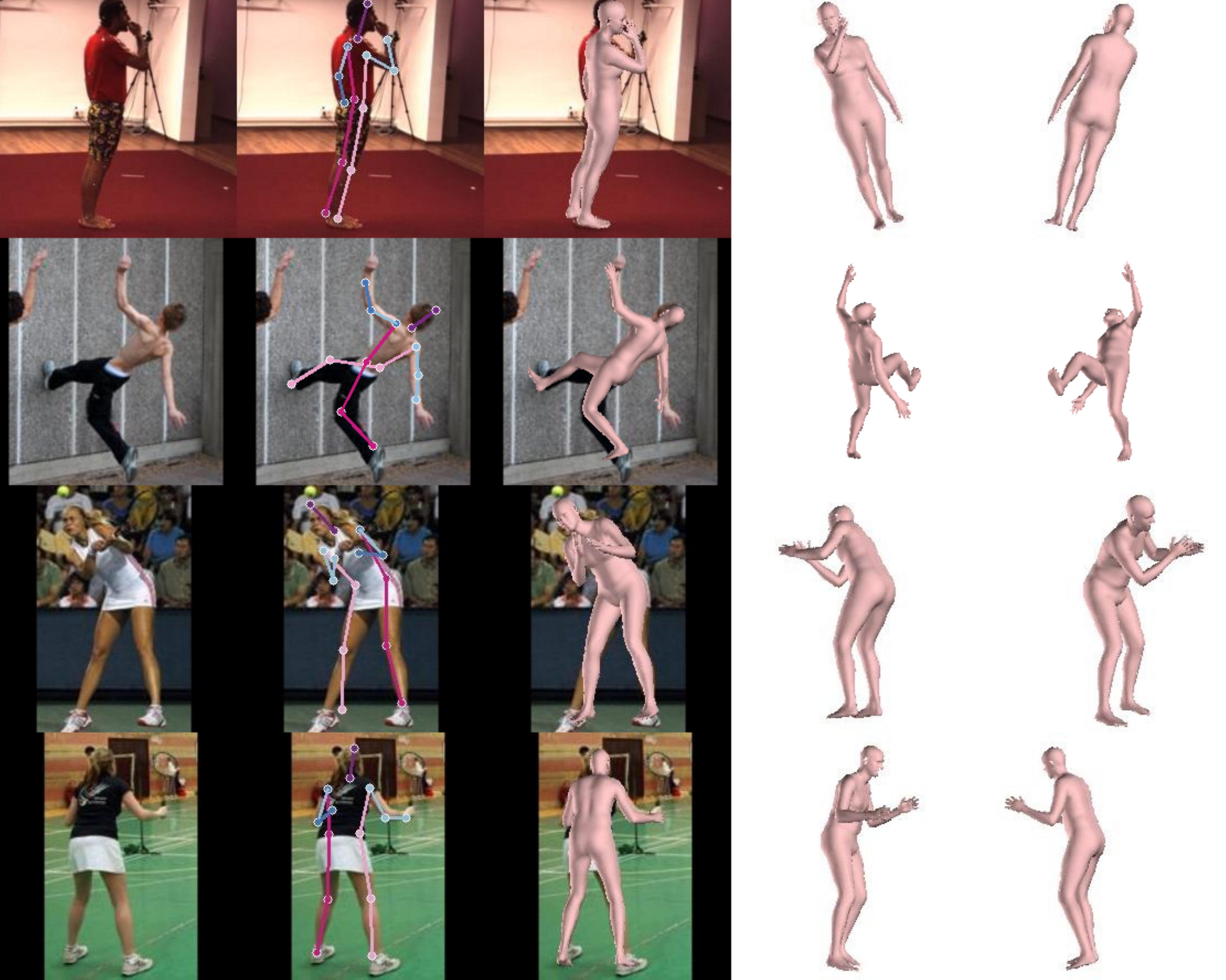}
    \caption{Qualitative SMPL parametric results. From left to right: input image, pose and shape, two different view perspectives.}
    \label{fig:extra_qualitative_results_smpl}
\end{figure*}

\noindent\textbf{Qualitative comparison} On the other side, we note that the shape of the parametric (SMPL) prediction is smooth but the shape of the non-parametric prediction may exhibit non-smooth artifacts. To demonstrate it, we shows examples of the rendered mesh for non-parametric prediction in Figure~\ref{fig:extra_qualitative_results}, and the ones for SMPL predictions in Figure~\ref{fig:extra_qualitative_results_smpl} for the evaluation images from the Human 3.6M, UP-3D and LSP datasets. To avoid some of the noise artifacts, we can apply some surface constraints, like vertex normal loss, to smooth the predicted surface for non-parametric methods in the future.

\subsection*{Qualitative results on occlusion}
The body is always self-occluded in a single image. It is especially challenging to predict the occluded limbs because of their rich poses from a variety of activities. This requires the methods to deduce the missing limbs from the other visible parts. Our model explicitly embed the whole joints and mesh in the bi-layer structure, enabling to learn the occluded part in a data-driven way. This bi-layer structure further learn the interactions between the body joints and mesh vertices by the fusion graph and thus guide the dense mesh with the sparse joints, which is less challenging to learn from the data. In Figure~\ref{fig:extra_qualitative_results} and~\ref{fig:extra_qualitative_results_smpl}, We render the predicted meshes from different views and demonstrate that our methods can always learn the occlusion well from the data. We also show a failure case of local hands pose in Figure~\ref{fig:extra_qualitative_results_bad}. Although the shape and pose seem correct from the camera view of the image, when viewed from different angles, we can see that the hands are separated rather than joined.


\end{document}